\documentclass[lettersize,journal]{IEEEtran}
\usepackage{amsmath,amsfonts}
\usepackage{array}

\usepackage[caption=false, font=footnotesize, labelformat=empty]{subfig}
\usepackage{textcomp}
\usepackage{stfloats}
\usepackage{url}
\usepackage{verbatim}
\usepackage{graphicx}
\usepackage{cite}
\usepackage[most]{tcolorbox}
\tcbuselibrary{listings}

\newtcbtheorem[auto counter, number within=section]{mytheorem}{Theorem}{
    enhanced,
    colback=green!5,
    colframe=green!35!black,
    fonttitle=\bfseries,
    arc=4pt %
}{thm}

\newtcblisting{pycode}{
  listing only,
  colback=gray!8,
  colframe=gray!50!black,
  arc=4pt,
  boxrule=0.5pt,
  listing options={
    language=Python,
    basicstyle=\ttfamily\small,
    keywordstyle=\color{blue!60!black},
    commentstyle=\color{green!40!black},
    showstringspaces=false,
    breaklines=true
  }
}

\newtcolorbox{examplebox}{
  colback=blue!3,
  colframe=blue!50!black,
  arc=4pt,
  boxrule=0.5pt
}

\usepackage{amssymb} 

\usepackage[table]{xcolor}
\definecolor{Viri1Base}{HTML}{440154}
\definecolor{Viri2Base}{HTML}{3B528B}
\definecolor{Viri3Base}{HTML}{21918C}
\definecolor{Viri4Base}{HTML}{5EC962}
\definecolor{Viri5Base}{HTML}{FDE725}

\colorlet{colour1}{Viri1Base!25!white}
\colorlet{colour2}{Viri2Base!25!white}
\colorlet{colour3}{Viri3Base!25!white}
\colorlet{colour4}{Viri4Base!25!white}
\colorlet{colour5}{Viri5Base!25!white}

\usepackage{stackengine}

\usepackage[ruled,vlined]{algorithm2e}

\usepackage{stmaryrd}

\newcolumntype{C}[1]{>{\centering\arraybackslash}m{#1}}
\newcolumntype{R}[1]{>{\raggedright\arraybackslash}m{#1}}
\newcolumntype{L}[1]{>{\raggedleft\arraybackslash}m{#1}}

\makeatletter
\def\bstctlcite{\@ifnextchar[{\@bstctlcite}{\@bstctlcite[@auxout]}}
\def\@bstctlcite[#1]#2{\@bsphack
  \@for\@citeb:=#2\do{%
    \edef\@citeb{\expandafter\@firstofone\@citeb}%
    \if@filesw\immediate\write\csname #1\endcsname{\string\citation{\@citeb}}\fi}%
  \@esphack}
\makeatother

\usepackage{booktabs}
\usepackage{multirow}

\usepackage[T1]{fontenc}

\usepackage{calc} 

\usepackage[pdfstartview=XYZ,
bookmarks=true,
colorlinks=true,
linkcolor=blue,
urlcolor=blue,
citecolor=blue,
bookmarks=true,
linktocpage=true, 
hyperindex=true
]{hyperref}

\usepackage{orcidlink}

\begin{document}
\bstctlcite{IEEEexample:BSTcontrol} 

\title{On the Origin of Synthetic Information\\by Means of Steganographic Inheritance}

\author{Ching-Chun Chang and Isao Echizen

\thanks{C.-C. Chang and I. Echizen are with the Information and Society Research Division, National Institute of Informatics, Tokyo, Japan.
}
\thanks{Correspondence: C.-C. Chang (email: ccchang@nii.ac.jp)
}
}

\maketitle

\begin{abstract}
The origin of species has been the mystery of mysteries in natural science. By analogy, the origin of synthetic information, we suggest, is the mystery of mysteries in information science. The question carries a moral weight that a technical account can neither fully resolve nor responsibly ignore, as its impact on truth, trust, and human intellect extends deep into the broader economy and society. The very power of artificial intelligence makes the evolutionary lineage of synthetic information grow ever harder to trace, for a sufficiently capable model may generate offspring that bear little resemblance, at either the structural or signal level, to the parent source from which they were derived. As in genetics, two individuals may share the same phenotype mirroring each other in outward appearance, yet differ fundamentally in their genotype. We propose, by means of steganography, a mechanism analogous to heredity. At the moment an offspring is reproduced, a projector derives a trait from the parent, and a steganographic encoder invisibly hides it within the offspring. This trait persists throughout the offspring’s life cycle in a cyber ecosystem. When parentage is queried, a steganographic decoder extracts the trait from the offspring and compares it against the traits of candidate parents in a reference pool, thereby nominating the most likely one. A theoretical analysis characterises phylogenetic accuracy as a function of projector and stegosystem properties, whilst empirical evaluations across multiple projectors and stegosystems demonstrate the viability of the proposed methodology under a broad spectrum of processing operations and semantic modifications. We envision a cyber ecosystem in which synthetic information, endowed with hidden yet traceable lineage traits, branches from a simple beginning into endless forms that have been, and are being, evolved.
\end{abstract}

\section{Introduction}
\IEEEPARstart{T}{he} origin of species, remarked during the dawn of modern science as the mystery of mysteries, continues to captivate the human mind. The mystery Darwin sought to resolve was one that had long vexed naturalists~\cite{darwin1859}:
 
\begin{quote}
\emph{... until it could be shown how the innumerable species inhabiting this world have been modified, so as to acquire that perfection of structure and coadaptation which most justly excites our admiration.}
\end{quote}
The answer he arrived at, descent with modification through natural selection, was at once simple and world-altering.

We write at a moment when an analogous mystery presents itself in a new domain, one formed around the emergence of artificial intelligence (AI)~\cite{Turing:1950aa, Rumelhart:1986aa, LeCun:2015aa}. The advancement of AI has given rise to a prolific and ever-accelerating generation of synthetic information, in the form of imagery, audio and text, along with the manifold forms of digital content that now circulate through the cyber world at a scale and speed that preclude any comprehensive human oversight~\cite{Lazer:2018aa, Floridi:2018ac, Vosoughi:2018aa, Knott:2024aa}. Synthetic information does not arise from nothing. Like organisms in an ecosystem, pieces of synthetic information descend from other pieces. Generative models continuously reshape digital content, whether by weaving a patch of imagery into a scene, a snippet of voice into a speech, or a fragment of text into a document~\cite{DBLP:journals/corr/KingmaW13, NIPS2014_5ca3e9b1, NEURIPS2020_4c5bcfec, NEURIPS2020_1457c0d6}. The understanding of this evolution requires tracing its lineage back to its roots, uncovering its provenance, its progenitor, and the chain of generation that produced it. To ask from which source any given piece of synthetic information originates is, we suggest, the mystery of mysteries in the information age.

The stakes of this mystery are not a matter of abstract concern. The question impacts truth, trust and the human intellect everywhere from our courtrooms and newsrooms to the creative industries, and well into the broader economy and society~\cite{BrundageEtAl2018, Chesney:2019aa, doi:10.1177/2056305120903408, 10183726}. In legal proceedings, a piece of fabricated evidence can decisively alter a verdict; in journalism, a reporter who publishes a story without verifying its origin may unwittingly spread misinformation; and in creative industries, a creator's work may be used without consent as the source material for a derivative piece. When synthetic information is used in the commission of harm, to trace its genealogy is to trace the architecture of the harm itself. The question of origin carries a moral weight that no technical account can fully discharge but that no technical account can responsibly ignore.

\begin{figure}[!t]
\centering
\includegraphics[width=1.0\linewidth]{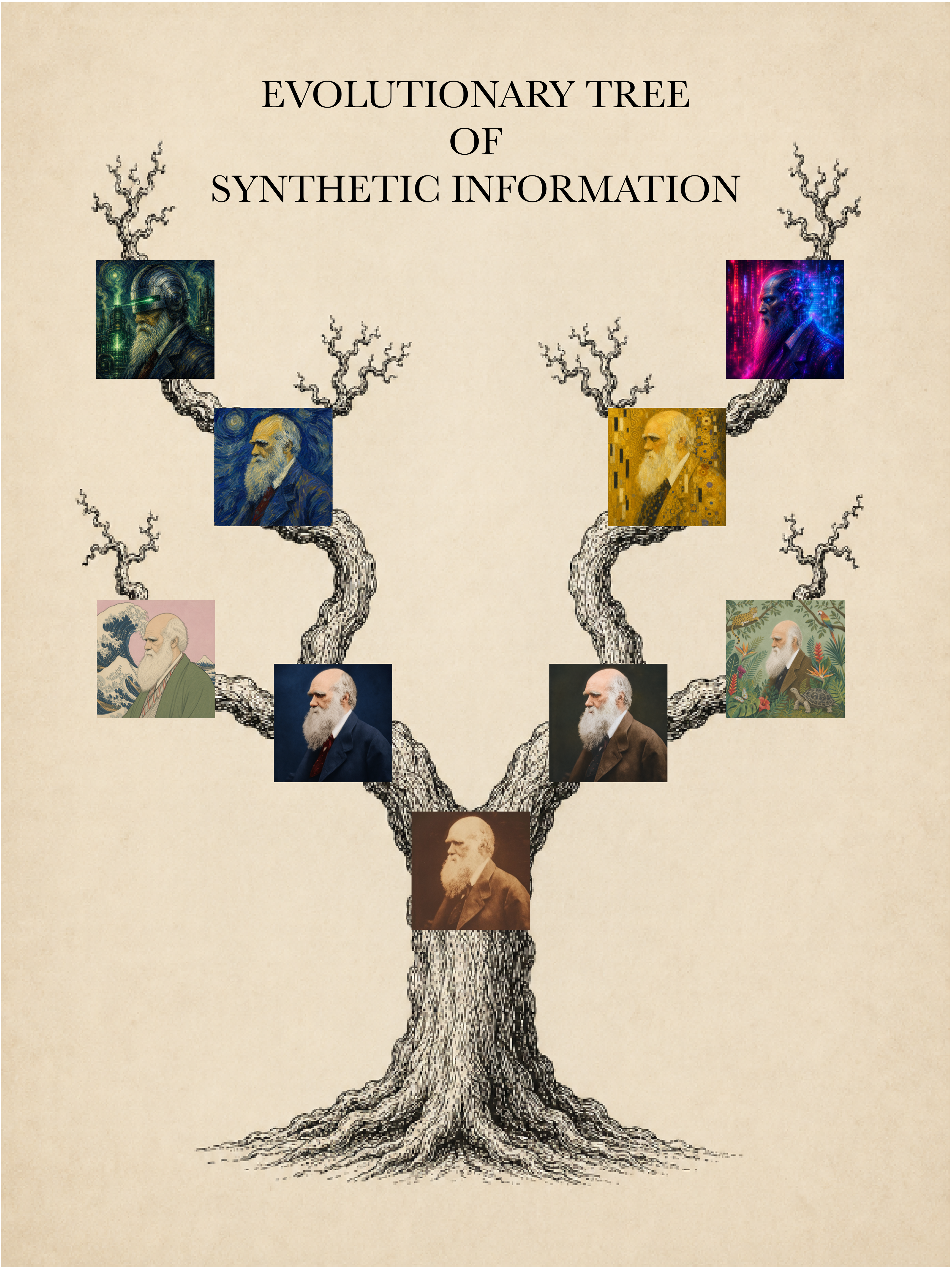}
\caption{Evolutionary tree of synthetic information, depicting the phylogenetic divergence of descendants from ancestral sources across generations.}
\label{fig:tree}
\end{figure}

It might seem that the tracing of such origins is an intuitive process. If a piece of synthetic information is derived from another, one may expect them to resemble one another, and that ensuing resemblance would, in principle, reveal the relation. This expectation underlies the forensic methods that have long dominated the field~\cite{6497035}. Yet this assumption might warrant closer examination under evolving generative paradigms. A sufficiently capable generative model may produce offspring that bear little resemblance, at either the structural or signal level, to the parent source from which they were derived. Consider, for example, cases where a model regenerates an image to recontextualise a focal subject within an entirely foreign narrative, resulting in a thorough overhaul of its physical state, aesthetic tone and atmospheric mood, as illustrated in Figure~\ref{fig:tree}. This introduces a profound predicament, where the very power of generative models makes the evolutionary lineage of synthetic information grow ever harder to trace to its origin.

In the natural world, surface appearance alone may be an unreliable proxy for phylogenetic lineage. Living organisms that resemble one another may carry a substantially different biological inheritance. As Schrödinger observed in \emph{What is Life}~\cite{Schrodinger:1944aa}: 
\begin{quote}
	\emph{The fact that two individuals may be exactly alike in their outward appearance, yet differ in their inheritance, is so important that an exact differentiation is desirable. The geneticist says they have the same phenotype, but different genotype.}
\end{quote}
At the very moment of reproduction, the lineage of an organism is encoded in its offspring as a compact and faithful record of ancestry, persisting throughout its life cycle. We seek, by means of steganography, a mechanism analogous to heredity for tracing the lineage of evolving synthetic information. At the moment of generation, one would embed the trait of the parent into its offspring, and, at the point of verification, extract it to retrieve the parent-offspring relationship, rather than attempting to infer relations from superficial resemblance. The trait, once embedded, should persist throughout the content's life cycle, surviving a reasonable range of modifications and transformations as it circulates through the cyber ecosystem in the wild. We call this steganographic inheritance, which constitutes the central proposition of this study.

\section{Preliminaries}
We organise the relevant literature by the nature of the question being asked. Any enquiry into a piece of media may be understood as an instance of one of the fundamental questions: what the media contains, whether it bears certain properties, where and when it came into being, how and why it was created, and from which source or whom it originates~\cite{10.1145/1978802.1978805}. Each of these questions has given rise to a body of work in its own right, and each illuminates a different facet of the media and the world that gave rise to it. We survey them in turn, before situating this study therein.

\subsection{What}
What information does the query carry? This is perhaps the most fundamental question one can ask about any media. Answers may be absolute, describing the content of the query in itself, or relative, characterising the query in relation to other media.

The absolute perspective is exemplified by recognition and description tasks, where the goal is to identify what objects, scenes, or concepts a query depicts~\cite{726791, 10.1007/978-3-642-15561-1_2, 5487377, 1597116, 10.1162/tacl_a_00177, 7298935}. The relative perspective is exemplified by retrieval and clustering tasks, where the goal is to find what other media the query resembles or shares content with~\cite{1238663, 4270197, 5235143, 10.1007/978-3-319-10590-1_38, 10.1007/978-3-319-46466-4_15}.

A further question concerns what information the query once carried. This may be addressed by self-recovery methods that embed within content, at the moment of its creation, the information needed to restore itself should it later be damaged or corrupted~\cite{817228, 6355682, 10238689}.

\subsection{Whether}
Whether a certain property holds for the query? This question underlies much of what we call media forensics~\cite{4806202, Wu:2015aa, 9010912, 9115874, 10.1145/3425780, 10.1162/coli_a_00549}. Such properties may be intrinsic, inherent to the query by virtue of how it was originally produced, or extrinsic, acquired through subsequent operations applied to the query after its creation.

The intrinsic perspective is exemplified by the detection of fully synthetic content, where the question is whether a query originates from a generative process rather than physical reality~\cite{1381784, 9156876, 10095167}. The extrinsic perspective is exemplified by the detection of manipulation and tampering, where the question is whether a query has been altered from some prior authentic state~\cite{723401, 1381775, 8578214, 9157215, 9753668}.

\subsection{Where/When}
Where or when was the query created? These questions situate the query in physical space and time.

In practice, such questions are often addressed through metadata and auxiliary information associated with the query at the time of its creation, including timestamps, geolocation tags, and device-recorded contextual data~\cite{10.1145/359340.359342, Haber:1991aa, 267415, 5732683}. When such metadata is absent or untrusted, spatial and temporal context may instead be inferred directly from the media content itself~\cite{4408995, 4587784, 10.1145/1526709.1526812, Lalonde:2010aa, 10.1007/978-3-319-61204-1_6}. The recovery and authentication of such spatiotemporal evidence carries implications across multiple domains. In legal proceedings, it may be decisive in corroborating or contesting an alibi, or in reconstructing a timeline of events. In journalism, it bears on disputes surrounding the true origin of media.

\subsection{How/Why}
How was the query produced or processed, and why? These questions seek to trace what gave rise to the media. The two are treated together here, for the pursuit of one naturally invites the other.

The question of how concerns the recovery of generative processes and their underlying parameters. A piece of media may be the product of sequential operations. Each operation may, to varying degrees, leave behind identifiable traces. Such traces are not merely a matter of identifying what occurred, but of inferring the parameters under which it occurred, including the adjustment factors applied in the processing of media~\cite{771070, 951529, 5487389, doi:10.2352/ISSN.2470-1173.2018.07.MWSF-213, 11370185}, or the language instructions that shaped a synthetic output~\cite{morris2024language, 10.5555/3698900.3699226, 10658265}.

The question of why is more elusive. To ask why a query was created is to ask after the intent of its creator. This is a question beyond the bounds of computing. In literary theory and aesthetics, the \emph{intentional fallacy} cautions that a work, once produced, stands independent of its authorial intent, and that intent cannot be reliably interpreted from the work itself~\cite{Wimsatt:1946aa}. In legal context, the long-standing debate between originalism and textualism turns on precisely this tension of whether one ought to seek the original intentions behind a text, or confine oneself to what the text itself permits~\cite{Powell:1985aa}. Even granting that intent is discernible, it may not be singular. Insights from social psychology suggest that human motives behind any act are often multifaceted, partially non-conscious, and resistant to precise articulation~\cite{Nisbett:1977aa}.

We do not claim that knowing how entails knowing why, nor that it is, in general, a well-posed question to ask for why in a computing sense as yet conceived. We raise it here to acknowledge that such traces tell, however imperfectly, of something beyond mere process.

\subsection{Who/Which}
Which source did the query originate from? Who created it? And through what chain of descent has it arrived at its present form? These questions lie at the heart of provenance.

Sources may be physical or digital in nature. In the case of physical sources, the question may ask which capture device produced the media~\cite{1418853, 1634362, 5439866, 8713484}. In the case of digital sources, the question may ask which generative model produced it~\cite{8695364, uchendu-etal-2020-authorship, uchendu-etal-2021-turingbench-benchmark, pmlr-v202-kirchenbauer23a}. In matters of authorship and ownership, the question concerns the identity of the creator, rights holder or responsible recipient~\cite{650120, 687830, 771066, 771072, 841169, 7555393}. One may ask further after the phylogenetic relation of the query to other media~\cite{BMVC.22.50, 10.1145/1459359.1459406, 6858004, 7118711, 8296535}. Such a relation may be undirected, indicating the existence of a connection without determining the hierarchy among them, or directed, seeking to identify the ancestral origin from which the query descends.

\subsection{Research Gap}
The questions surveyed above are not equally well-posed, nor equally well-answered. What unites them is a shared ambition to reason about something that lies beyond the observed media. This study concerns itself with directed phylogenetic analysis of synthetic media.

This question has been approached in the literature through passive means, typically by searching for similar instances and assessing how well one can explain or reconstruct another~\cite{5711452, 6030928, 8438504}. Prior art finds its natural domain in media that has undergone conventional operations. Under such conditions, there is reason to suppose that similarity at the level of signal or structure bears faithfully upon lineage, and that the asymmetry of explainability affords some grounds for inferring the causal direction of derivation.

When confronting modern generative processes, however, both assumptions invite a rethink. A derivative may inherit particular aspects of its source whilst bearing little resemblance to it in terms of signal or structure. Suppose two photographs of the same individual are taken independently, and that the first is used to generate synthetic content in which only the identity of the subject is preserved, while the expression, pose, apparel, and background are altogether altered. The derivative bears a relation to the first photograph and none to the second, yet all three are alike in terms of biometric identity. What it means for two pieces of media to be similar thus becomes a matter of some delicacy. The correspondence between explainability and causal direction is no less uncertain. A sufficiently capable generative model may reconstruct a parent from its offspring as readily as the reverse, or more so.

\begin{figure*}[!t]
\centering
\includegraphics[width=0.75\linewidth]{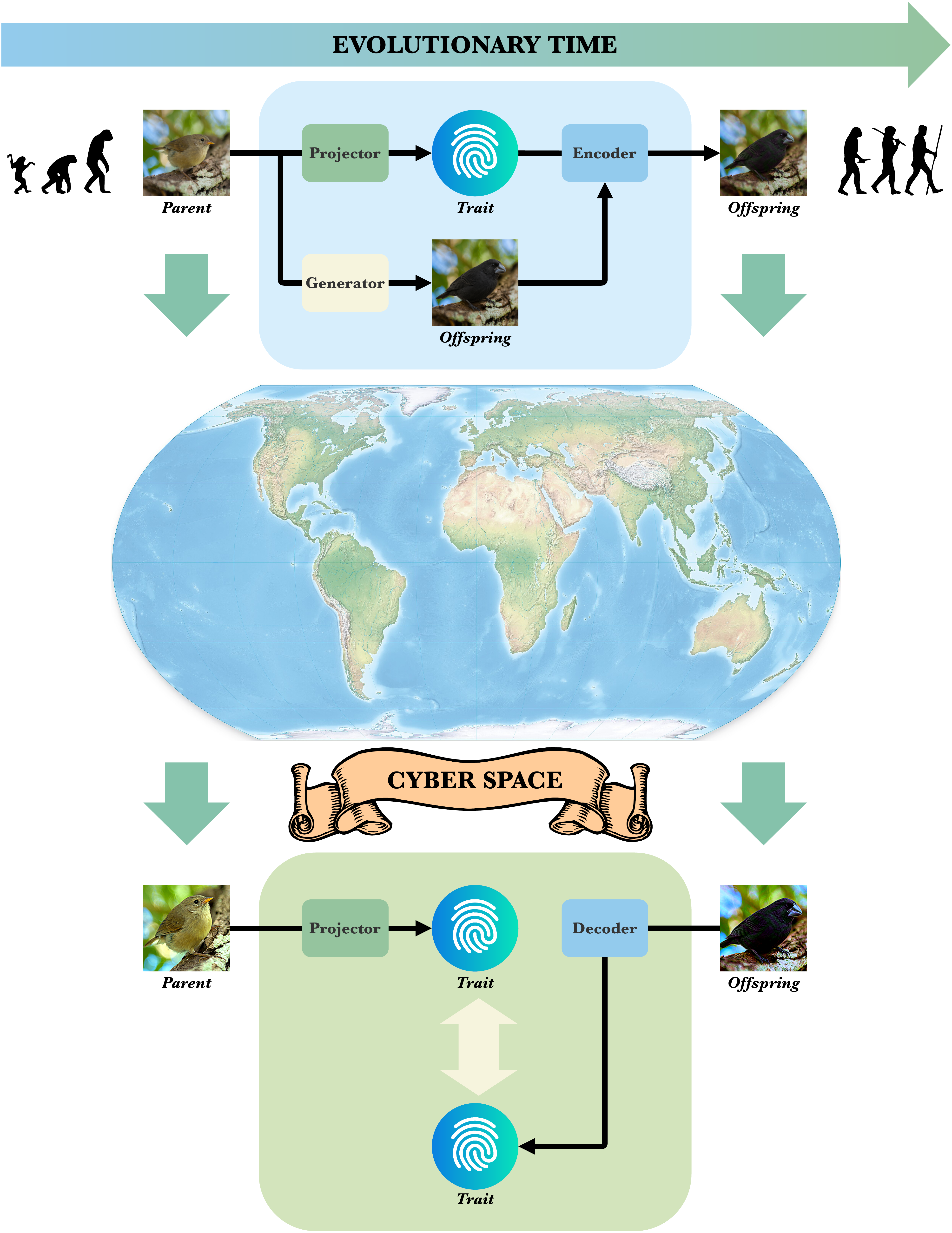}
\caption{Overview of steganographic inheritance across evolutionary time and cyber space. At each act of generation, a trait is derived from the parent and embedded within the offspring. Upon enquiry, a trait extracted from the query is compared against that derived from the candidate for identifying the parent-offspring pair.}
\label{fig:overview}
\end{figure*}

\section{Methodology}
The methodology that follows sets out the problem under consideration, the system devised to address it, and the theoretical framework that characterises its expected behaviour.

\subsection{Problem Statement}
Let a piece of synthetic content under investigation be referred to as a query and a finite collection of content assembled at the time of enquiry as a pool. The direct source from which the query was derived is referred to as its parent. The parent may or may not be present in the pool, alongside content that is ancestrally related or unrelated to it. Both the query and any item in the pool may have undergone arbitrary post-generation modifications prior to the moment of enquiry. A good solution should, with high probability, return the direct parent of the query if it is present in the pool, and a null response otherwise. Its merit lies in correctly identifying true genealogical links, withholding judgement in the absence of the parent, resisting false attribution in the presence of unrelated or superficially similar content, and ensuring that a broken link at any single node propagates no further along the lineage.

\subsection{System Overview}
We address this problem through a scheme of steganographic inheritance. The central idea is that, rather than inferring genealogical relationships passively from content, one may embed them actively at the moment of generation such that every piece of synthetic media carries, within itself, a faithful trace of its origin. We refer to the source as the \emph{parent} and to the generated counterpart as the \emph{offspring}. The system comprises two phases: a \emph{forward} phase carried out at the moment of generation, and a \emph{backward} phase carried out at the moment of retrieval, as illustrated in Figure~\ref{fig:overview}.

 In the forward phase, at each act of generation, a compact representation of the parent, referred to hereafter as a \emph{trait}, is extracted and embedded into the offspring by means of a steganographic encoder. The offspring thus inherits the trait from its parent, in a manner that persists throughout its life cycle, encompassing all subsequent handling, including storage, transmission and processing in cyber space. This evolutionary process may repeat across successive generations, each offspring in turn becoming a parent, carrying forward the lineage. 
 
 In the backward phase, given a query and a pool, a steganographic decoder extracts the trait embedded within the query. This extracted trait is then compared against the corresponding representation of each candidate in the pool. The candidate whose representation most closely agrees with the extracted trait is nominated. Should the degree of agreement fall below what may be considered a credible match, the nomination is withheld and the parent is deemed absent from the pool.

The scheme assumes the cooperation of the generative platform. It is only within a cooperative framework that inheritance takes place at each step of generation. What becomes of the media thereafter, whether it is further edited, redistributed or re-generated by other means, lies beyond the platform's control, and the system aspires to remain reliable in the face of such eventualities.

\subsection{Trait}
The trait is a compact binary representation of the content, taking the form of a vector in $\{0, 1\}^n$, where $n$ is a fixed length. This representation should be stable enough to serve as an identifier; that is, two copies of the same content, under typical post-generation handling, ought to yield representations that agree closely, whilst two copies of distinct content ought not. For digital images, the handling it may encounter in practice would include adjustments to light, colour, details and geometry.

In principle, any mapping from content to $\{0, 1\}^n$ may serve as the \textit{projector}. Practically, this mapping may be realised by cryptographic hashing, by perceptual hashing, or by an arbitrary feature extractor whose output is subsequently projected into the binary domain by means of a random projection. Specifically, in the case where the features are not of dimension $n$, a seeded pseudo-random matrix is used to project the feature vector into a representation of length $n$, whereupon the sign of each component yields the final $n$-bit trait. The use of a seeded matrix ensures that the same projection is applied consistently across both the generation and retrieval phases.

\begin{figure*}[!t]
\centering
\includegraphics[width=1.0\linewidth]{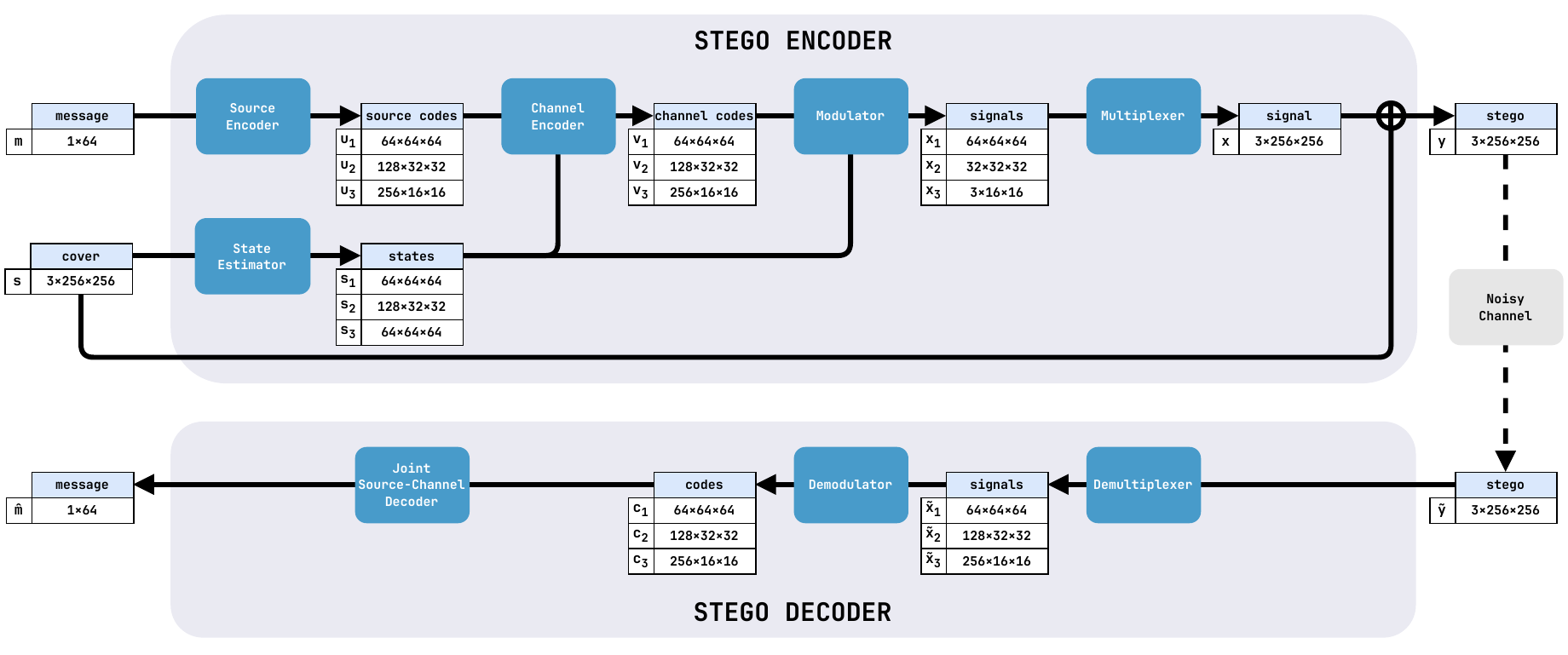}
\caption{General architecture of the steganographic system operating across multiple scales of resolution with communication-based modules.}
\label{fig:stegosystem_overview}
\end{figure*}

\begin{figure*}[!t]
\centering
\includegraphics[width=1.0\linewidth]{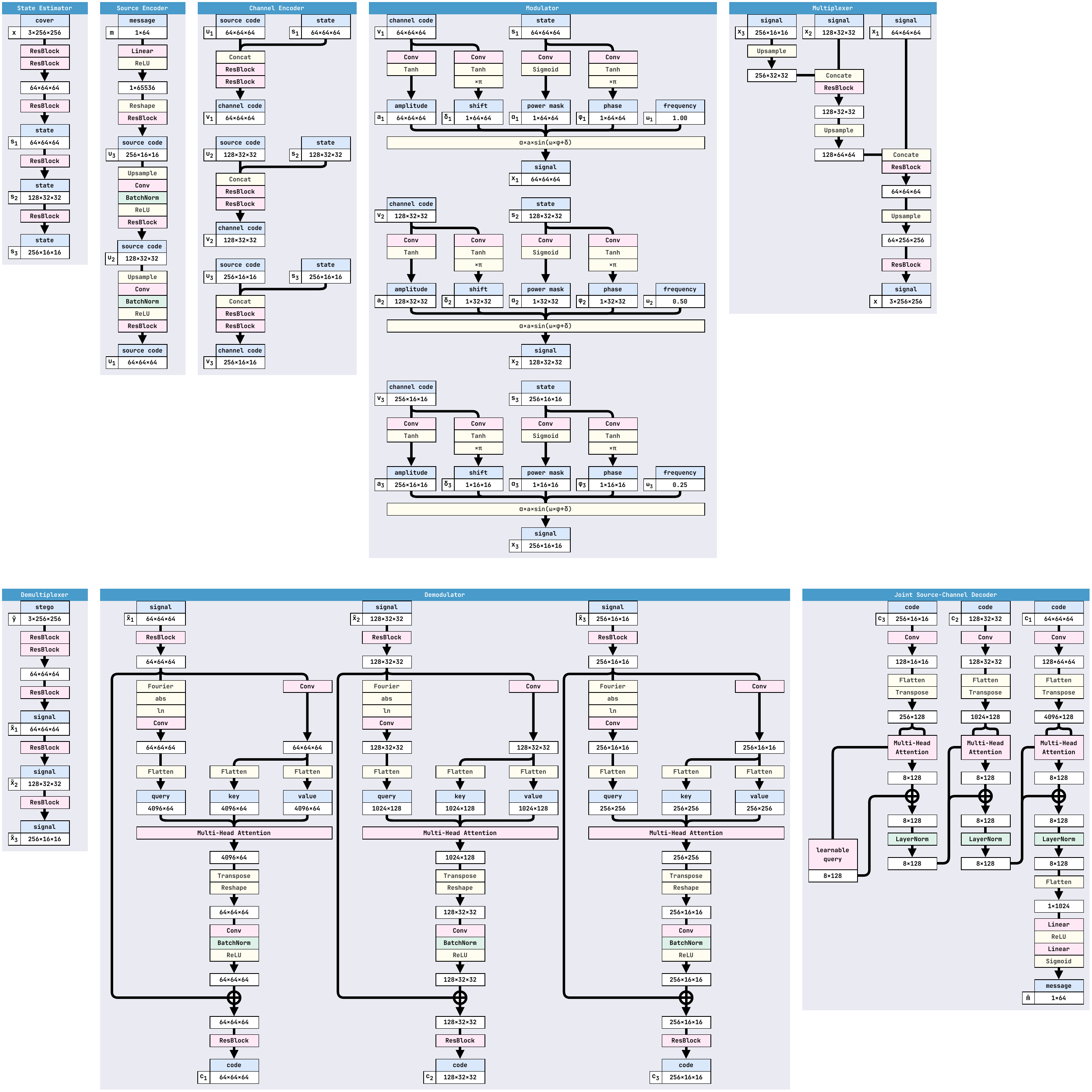}
\caption{Detailed neural network architecture of each module in the encoder and decoder.}
\label{fig:stegosystem_modules}
\end{figure*}

\subsection{Inheritance}
The steganographic system is cast in the form of info-communications. The problem of embedding a message into a cover in the presence of a known state, the carrier itself, is an instance of communications with side information available at the transmitter~\cite{1056659, 771068, 1184136, 1188745, 1298836}. In the standard formulation, the received signal is given by
\begin{equation}
	Y = X + S + Z ,
\end{equation}
where $X$ is the transmitted signal, $S$ is the known channel state, and $Z$ is the unknown channel noise (though its distribution may sometimes be anticipatable). The informed transmitter may exploit knowledge of the channel state to mitigate, and under idealised conditions to cancel, the interference. Complete cancellation is theoretically attainable, but is precluded in practice by the power constraint and the presence of channel noise. By analogy, the channel state is the cover, the transmitted signal is the embedded message, the power constraint corresponds to the perceptual fidelity constraint, and the channel noise corresponds to the processing of the media during storage and transmission.

The steganographic system is constituted by an encoder and a decoder, each operating at multiple scales to process signals across a hierarchy of resolutions, as illustrated in Figure~\ref{fig:stegosystem_overview}. The encoder comprises a state estimator, a source encoder, a channel encoder, a modulator and a multiplexer; the decoder comprises a demultiplexer, a demodulator and a joint source-channel decoder, as detailed in Figure~\ref{fig:stegosystem_modules}.

On the encoder side, the state estimator and source encoder extract multi-scale representations of the cover and the message, respectively. The channel encoder transforms the source codes into channel codes at each scale, conditioned on the corresponding cover state representation. The modulator converts the channel codes into signals, each taking the form of a sinusoidal carrier, or simple harmonic motion,
\begin{equation}
	\alpha \cdot a \cdot \sin(\omega \cdot \phi + \delta) ,
\end{equation}
where $\alpha$ and $\phi$ are derived from the state for power masking, or perceptual shaping in the steganographic sense, and phase control respectively, $a$ and $\delta$ are derived from the channel code as amplitude and shift, and $\omega$ is a constant frequency. The multiplexer aggregates the signals at all scales and the resulting signal is added to the cover to yield the stego.

On the decoder side, the demultiplexer separates the received stego into signals at each scale. The demodulator first computes the log-magnitude spectrum of the features via Fourier transform, and then fuses the resulting frequency features with the spatial features by means of a multi-head attention mechanism, in which the frequency features serve as queries and the spatial features serve as keys and values. The joint source-channel decoder maintains a set of learnable global queries, which iteratively aggregate multi-scale features by means of a multi-head attention mechanism from coarse to fine, and the resulting representation is decoded by a multi-layer perceptron into the message.

The stegosystem is trained to jointly optimise two objectives: fidelity, requiring the stego to remain perceptually close to the cover, and accuracy, requiring the message extracted by the decoder to agree with the message embedded by the encoder. Fidelity is measured by the $\ell_2$ and $\ell_\infty$ distances between the stego and the cover, whereas accuracy is measured by the binary cross-entropy between the extracted and embedded messages.

We term this steganographic system \emph{cognitive harmonic artificial steganographer} (CHAS), in which cognitive in reference to the state-aware transmitter; harmonic in reference to the sinusoidal form of the modulation; artificial in reference to the communication-like artificial neural network modules; and steganographer in reference to the steganographic system itself.

\begin{figure*}[!t]
\centering
\includegraphics[width=1.0\linewidth]{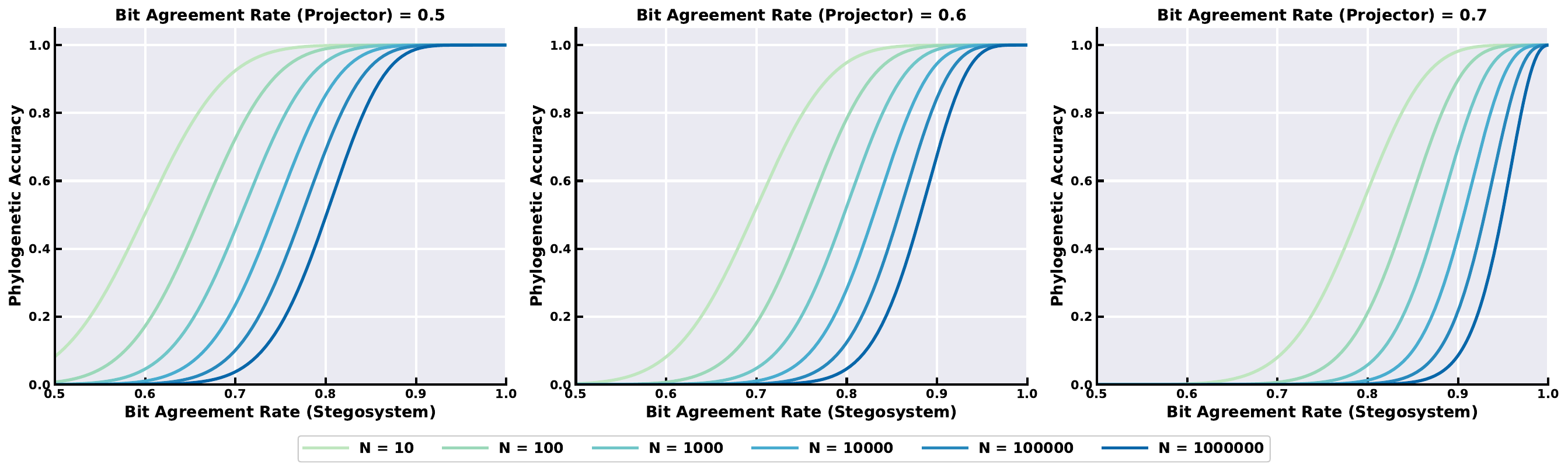}
\caption{ Theoretical phylogenetic accuracy as a function of stegosystem bit agreement rate, across varying projector bit agreement rates and pool sizes.}
\label{fig:plot1_retrieval_accuracy_panel}
\end{figure*}

\begin{figure*}[!t]
\centering
\includegraphics[width=1.0\linewidth]{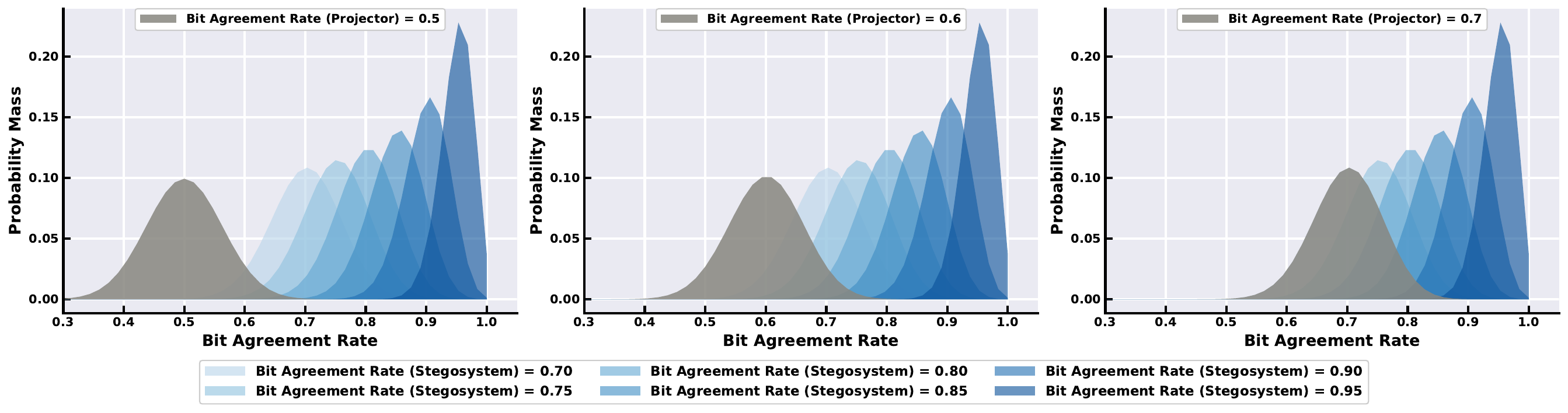}
\caption{Probability distributions of bit agreement rates for the true parent and unrelated candidates.}
\label{fig:plot2_distribution_overlap_panel}
\end{figure*}

\subsection{Theoretical Analysis}
Given a projector, a stegosystem and a pool of size $N$, one may wonder what the phylogenetic accuracy would be in the theoretical sense? In other words, it is a question about how probable the true parent is returned as the top match within the pool.

The performance of a projector and a stegosystem may be quantified by the bit agreement rate of the projector and the bit agreement rate of the stegosystem. The former refers to the probability that any single bit of the trait projected from a given piece of content agrees with the corresponding bit of the trait projected from an arbitrary, unrelated piece of content. The latter refers to the probability that any single bit of the trait extracted by the steganographic decoder agrees with the bit that was embedded by the steganographic encoder. An ideal projector's bit agreement rate is 0.5, indicating that the trait projected from one piece of content is statistically independent of that projected from any other, regardless of any superficial resemblance between them. An ideal stegosystem's bit agreement rate is 1.0, indicating that the trait is recovered without error, regardless of the processing the content may have undergone.

For the purposes of this analysis, we suppose the true parent to be present in the pool, and set aside the question of abstention; that is, the interest here is in which candidate ranks highest, rather than whether any candidate is sufficiently credible to be nominated. The phylogenetic accuracy is thus the probability that the true parent ranks highest among all $N$ candidates. Specifically, let $k$ denote the number of bits, out of $n$, in which the extracted trait agrees with the true parent's trait. The phylogenetic accuracy is then the probability that this count exceeds that of every unrelated candidate in the pool, as given by
\begin{equation}
\sum_{k=0}^{n} \mathbb{P}(H^\ast = k) \cdot \mathbb{P}(H < k)^{N-1} ,
\end{equation}
where $H^\ast$ denotes the bit agreement rate (Hamming similarity) between the trait extracted from the offspring and the trait of the true parent, and $H$ denotes the bit agreement rate (Hamming similarity) between the extracted trait and the trait of any unrelated candidate. The former follows a binomial distribution with each bit agreeing with probability $q$, equal to the bit accuracy; the latter follows a binomial distribution with each bit agreeing with probability $p$, equal to the bit agreement probability. Upon substituting the binomial expressions, the phylogenetic accuracy becomes
\begin{equation}
\sum_{k=0}^{n} \binom{n}{k} q^k (1-q)^{n-k} \cdot \left[\sum_{j=0}^{k-1} \binom{n}{j} p^j (1-p)^{n-j}\right]^{N-1} .
\end{equation}
Figure~\ref{fig:plot1_retrieval_accuracy_panel} plots the retrieval accuracy as a function of bit accuracy, for varying bit agreement probabilities and pool sizes $N$. Figure~\ref{fig:plot2_distribution_overlap_panel} illustrates the underlying separation between the two binomial distributions that gives rise to it.

\begin{figure*}[!t]
\centering
\includegraphics[width=1.0\linewidth]{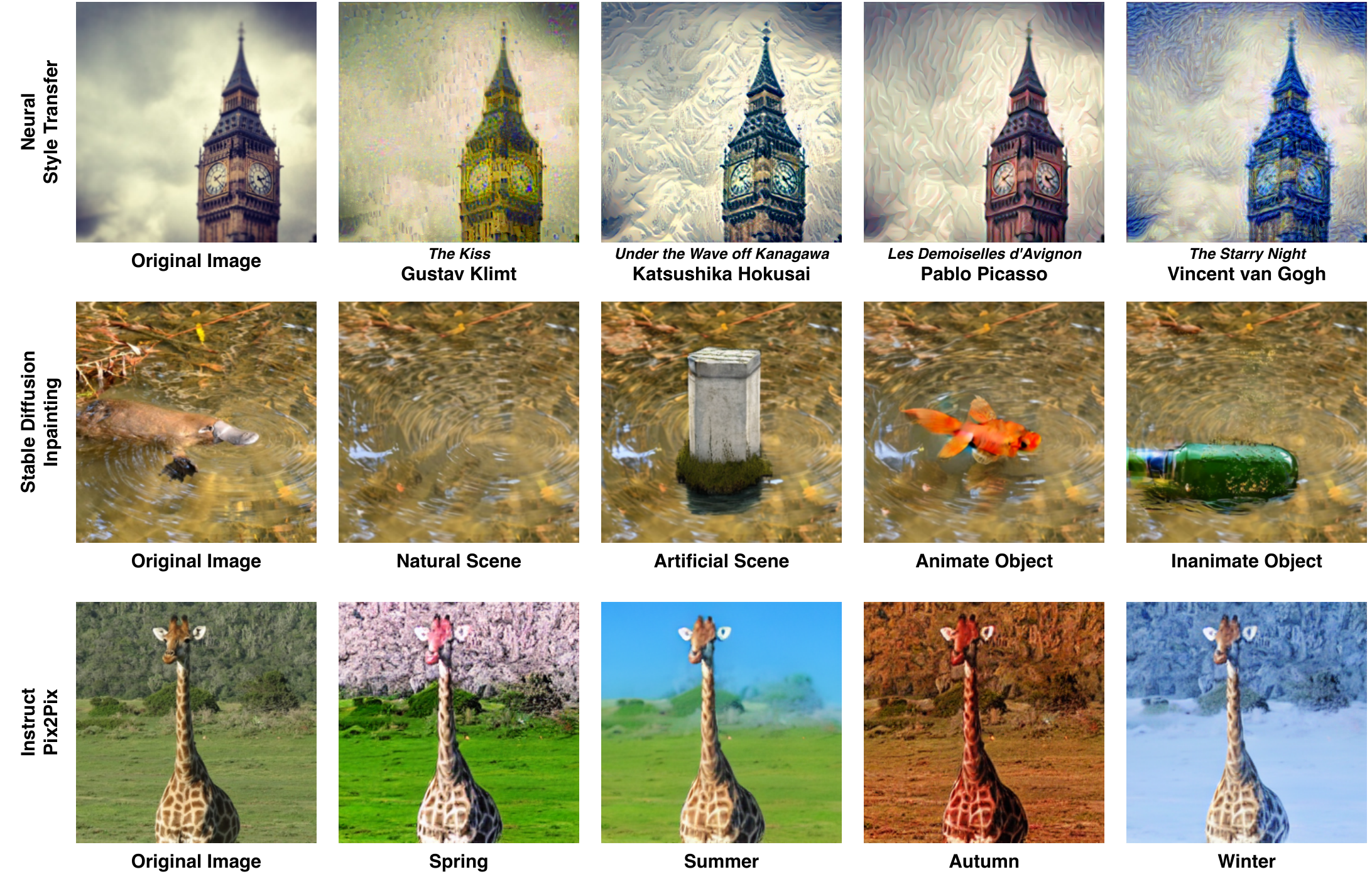}
\caption{Examples of semantic edits spanning stylistic, local and global transformations.}
\label{fig:semantic_edits}
\end{figure*}

\section{Experimental Setup}
The experimental setup that follows specifies the components under evaluation, the conditions to which media may be subjected, and the data on which the evaluation is conducted.

\subsection{Projectors}
Five projectors are evaluated, spanning three broad categories. The first is a cryptographic hash function SHA-256~\cite{Damgard:1989aa, 10.5555/118209.118249, 10.1007/11535218_2}. The second is a perceptual hash function pHash~\cite{899541}. The remaining three are neural feature extractors: ResNet~\cite{7780459}, CLIP~\cite{pmlr-v139-radford21a}, and DINO~\cite{9709990}. Of these, ResNet is based on a convolutional neural network~\cite{NIPS2012_c399862d}, optimised via supervised learning, whilst CLIP and DINO are based on vision transformers~\cite{Dosovitskiy:2021aa}, optimised via multimodal learning and self-supervised learning, respectively.

\subsection{Stegosystems}
Five stegosystems are evaluated. The first two are classical signal-processing methods: quantisation index modulation (QIM)~\cite{923725} and improved spread spectrum (ISS)~\cite{1188737}. The remaining three are learning-based methods: HiDDeN~\cite{Zhu:2018aa}, StegaStamp~\cite{Tancik:2020aa} and the proposed CHAS. All stegosystems are configured to embed a 64-bit message where possible. As HiDDeN and StegaStamp are pre-trained models, their capacities are fixed at 30 and 56 bits, respectively.

Both QIM and ISS operate in the discrete cosine transform (DCT) domain of the luminance channel. At the encoder, a set of seeded pseudo-random vectors is constructed, with one vector per embedded bit. The selected DCT coefficients are projected onto each vector to yield a scalar, which is then modified to encode the desired bit. In particular, QIM moves the projection to the nearest point of a dithered quantisation lattice assigned to the bit, whilst ISS adds a linear displacement whose sign encodes that bit. The difference between the modified and original projection is back-projected along the same vector into the DCT domain and the back-projections from all bits are added to the selected coefficients. At the decoder, each bit is extracted by projecting the same selected DCT coefficients onto the same vector, QIM decides by minimum distance from the scalar to the two quantisation cosets, and ISS decides by the sign of that scalar.

For the learning-based methods, HiDDeN represents an early development of neural stegosystems, whereas StegaStamp incorporates random transformations during training to improve robustness, with a spatial transformer network to correct geometric distortions and an error correction code to repair corrupted message bits. The proposed CHAS is also learning-based and trained with random transformations in terms of light, colour, details and geometry.

\subsection{Common Processing Operations}
Four categories of common processing are considered: light, colour, details and geometry~\cite{5995413, 10.1145/2790296, 10.1145/3181974, 10.1145/3550454.3555526}. The light category covers adjustments to brightness, contrast, and exposure. The colour category covers adjustments to saturation, warmth and tint. The details category covers adjustments to blurriness, sharpness and grain, as well as JPEG compression. The geometry category covers cropping, rotation, and horizontal and vertical perspective. In total, 14 operations are examined, spanning the range of transformations that digital image content may ordinarily undergo.

\begin{table*}[t]
\centering
\caption{Stegosystem Performance}
\label{tab:stegosystem_performance}
\renewcommand{\arraystretch}{1.2}
\setlength{\tabcolsep}{6pt}

\begin{tabular}{l *{6}{>{\centering\arraybackslash}p{0.125\linewidth}}}
\hline
\textbf{Stegosystem} &
\textbf{Capacity} $\uparrow$ &
\textbf{Accuracy} $\uparrow$ &
\textbf{PSNR} $\uparrow$ &
\textbf{SSIM} $\uparrow$ &
\textbf{VIF} $\uparrow$ &
\textbf{LPIPS} $\downarrow$ \\
\hline

QIM        & 64 bits & 0.9994 $\pm$ 0.0062 & 40.03 $\pm$ 0.55 & 0.9605 $\pm$ 0.0171 & 0.9989 $\pm$ 0.0026 & 0.0318 $\pm$ 0.0344 \\
ISS        & 64 bits & 1.0000 $\pm$ 0.0000 & 41.00 $\pm$ 0.62 & 0.9684 $\pm$ 0.0139 & 0.9974 $\pm$ 0.0024 & 0.0250 $\pm$ 0.0288 \\
HiDDeN     & 30 bits & 0.9970 $\pm$ 0.0241 & 34.04 $\pm$ 1.90 & 0.9502 $\pm$ 0.0331 & 0.9485 $\pm$ 0.0243 & 0.0233 $\pm$ 0.0300 \\
StegaStamp & 56 bits & 0.9996 $\pm$ 0.0025 & 30.20 $\pm$ 1.62 & 0.9139 $\pm$ 0.0457 & 0.8527 $\pm$ 0.0224 & 0.0288 $\pm$ 0.0139 \\
CHAS       & 64 bits & 1.0000 $\pm$ 0.0000 & 34.29 $\pm$ 0.36 & 0.9492 $\pm$ 0.0348 & 0.9974 $\pm$ 0.0068 & 0.1005 $\pm$ 0.0630 \\

\hline
\end{tabular}
\end{table*}

\begin{table}[t]
\centering
\caption{Stegosystem Complexity}
\label{tab:stegosystem_complexity}
\renewcommand{\arraystretch}{1.2}
\setlength{\tabcolsep}{6pt}

\begin{tabular}{l *{3}{>{\raggedleft\arraybackslash}p{0.2\linewidth}}}
\hline
\textbf{Stegosystem} &
\textbf{Model Size} &
\textbf{Encoder Size} &
\textbf{Decoder Size}
\\
\hline

QIM        & \textemdash & \textemdash & \textemdash \\
ISS        & \textemdash & \textemdash & \textemdash \\
HiDDeN     & 411,903 & 169,347 & 242,556 \\
StegaStamp & 54,253,955 & 1,749,247 & 52,504,708 \\
CHAS       & 18,487,692 & 12,761,164 & 5,726,528 \\

\hline
\end{tabular}
\end{table}

\subsection{Semantic Edits}
Three generative models are employed to simulate semantic edits. Neural style transfer (NST) renders the content in the style of one of four reference artworks~\cite{7780634}: \emph{The Kiss} by Gustav Klimt, \emph{Under the Wave off Kanagawa} by Katsushika Hokusai, \emph{Les Demoiselles d'Avignon} by Pablo Picasso and \emph{The Starry Night} by Vincent van Gogh. Stable Diffusion inpainting replaces a local region of the content, conditioned on prompts drawn from four subject categories~\cite{9878449}: natural scene, artificial scene, animate object and inanimate object. Instruct-Pix2Pix applies a global edit, conditioned on prompts drawn from four seasonal themes~\cite{10204579}: spring, summer, autumn and winter. In total, 12 variants are examined, encompassing a diverse spectrum of stylistic, local and global transformations, as shown in Figure~\ref{fig:semantic_edits}.

\subsection{Data Generation}
Images are drawn from the COCO dataset~\cite{Lin:2014aa}, resized to 256 $\times$ 256 pixels. Starting from a pool of 100 root images, a phylogenetic tree is constructed to a depth of 3 generations, with no generator repeated along any chain. This gives three branches per node at the first generation, two at the second, and one at the third. At each generation, the parent's trait is projected and embedded into the offspring by means of steganographic inheritance. A total of 1,600 images constitute a standard pool. The effect of expanding or reducing this pool is examined in subsequent evaluations.

\begin{figure*}[t!]
    \centering
    \subfloat[Light: Brightness]{
        \includegraphics[width=0.22\linewidth]{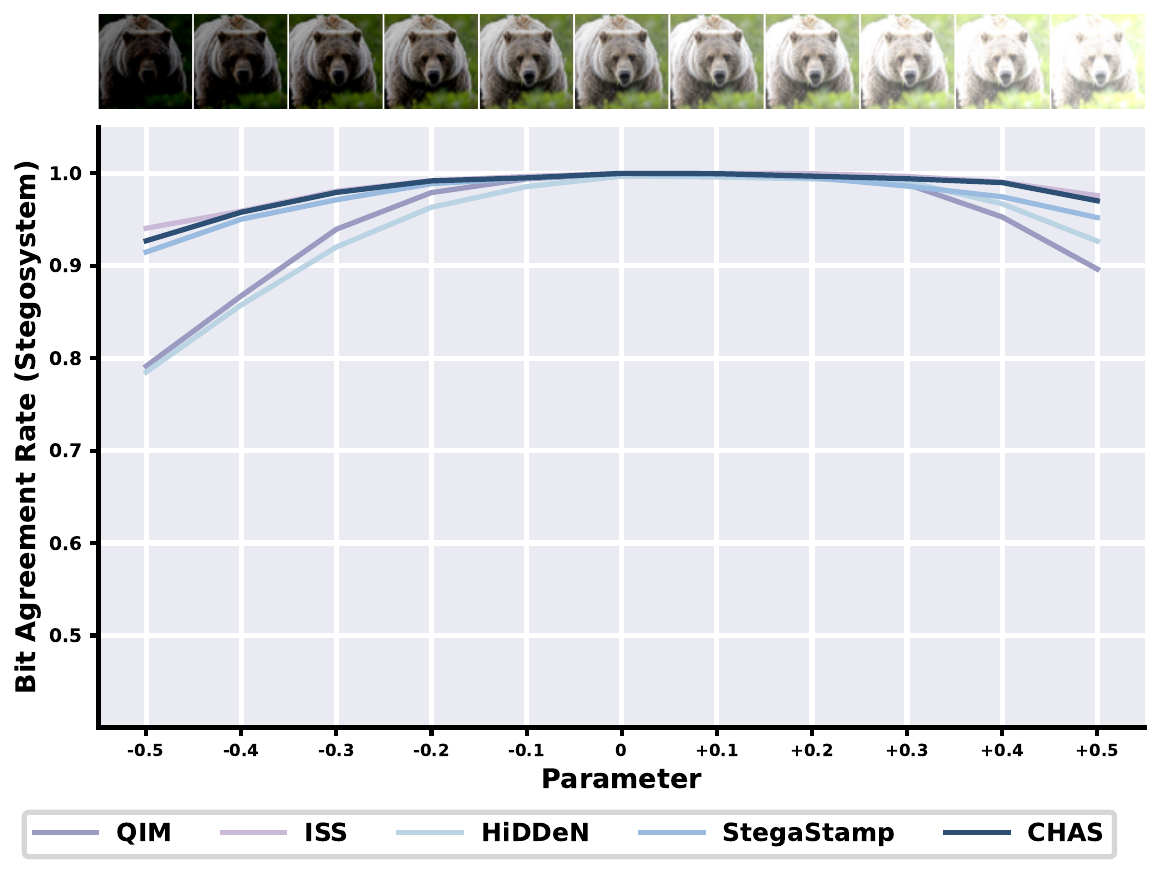}
    }
    \subfloat[Light: Contrast]{
        \includegraphics[width=0.22\linewidth]{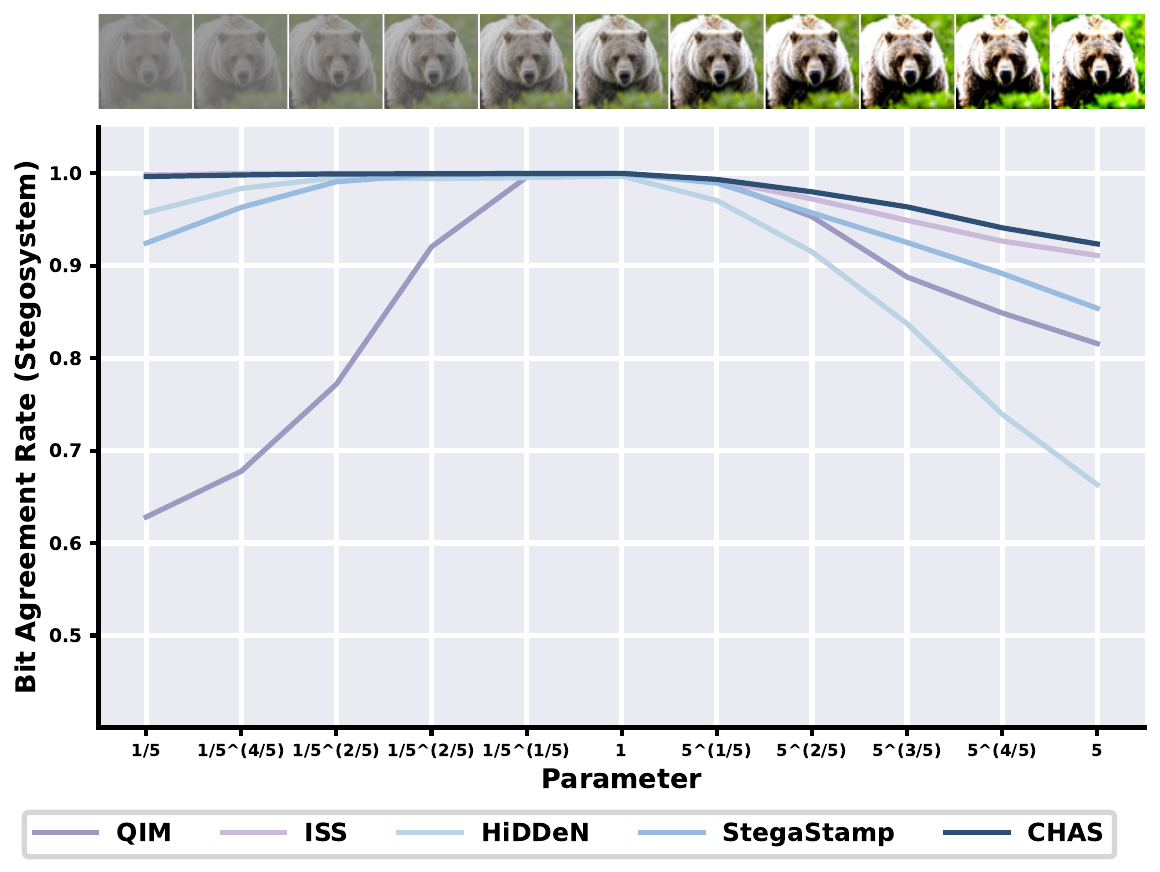}
    }
    \subfloat[Light: Exposure]{
        \includegraphics[width=0.22\linewidth]{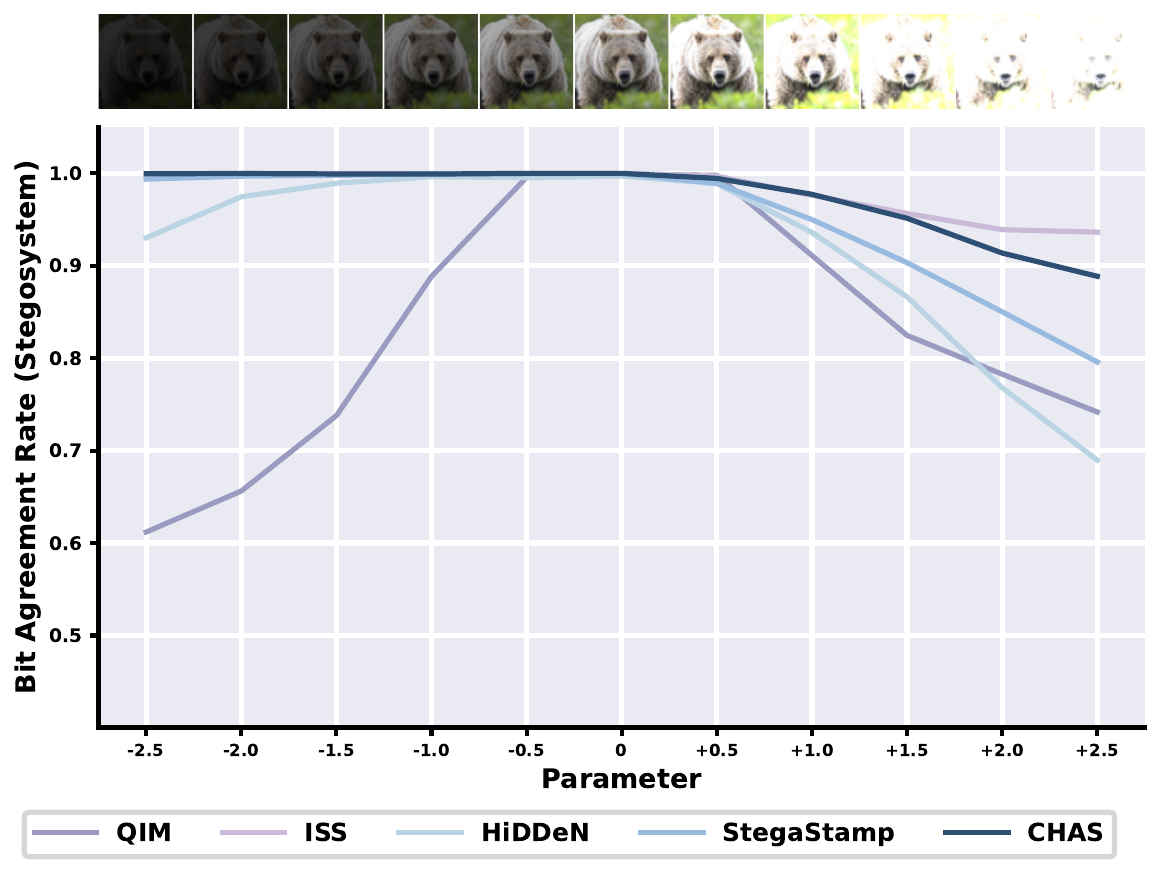}
    }
    \\
    \subfloat[Colour: Saturation]{
        \includegraphics[width=0.22\linewidth]{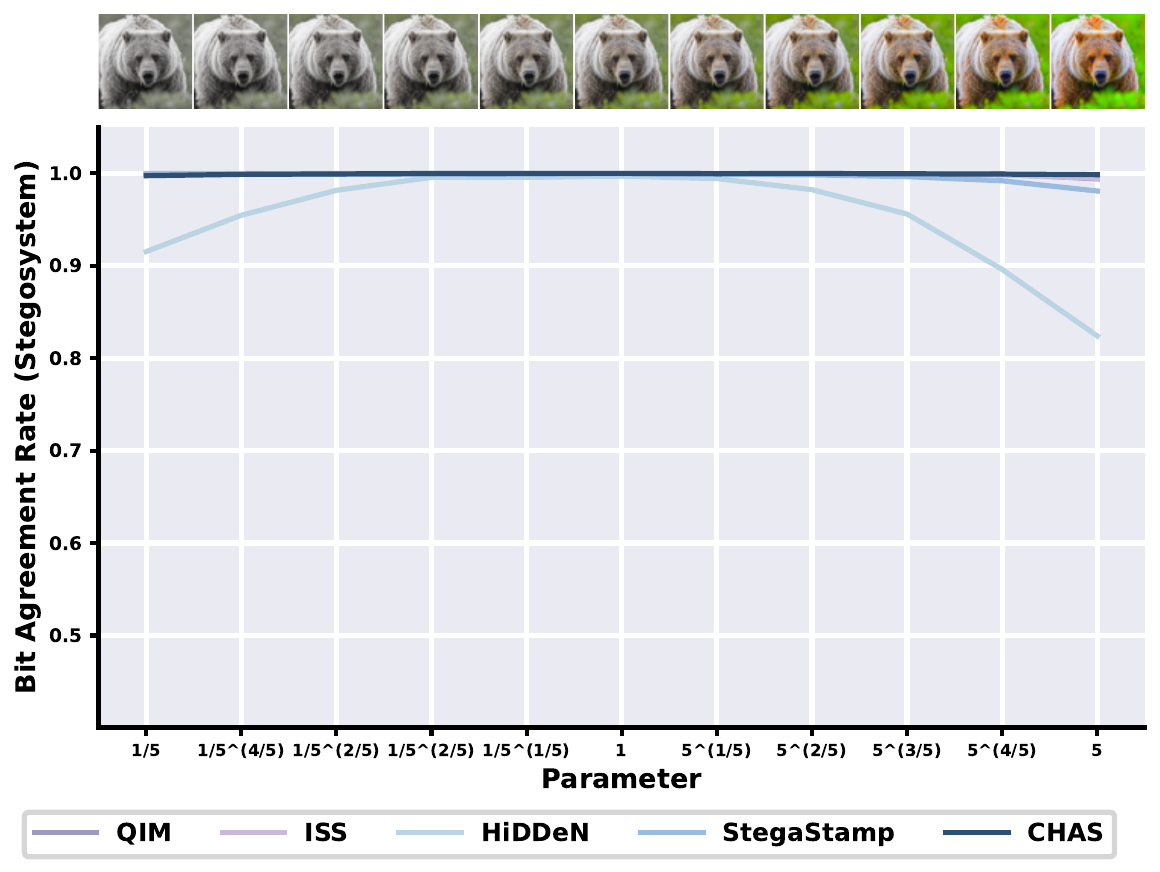}
    }
    \subfloat[Colour: Tint]{
        \includegraphics[width=0.22\linewidth]{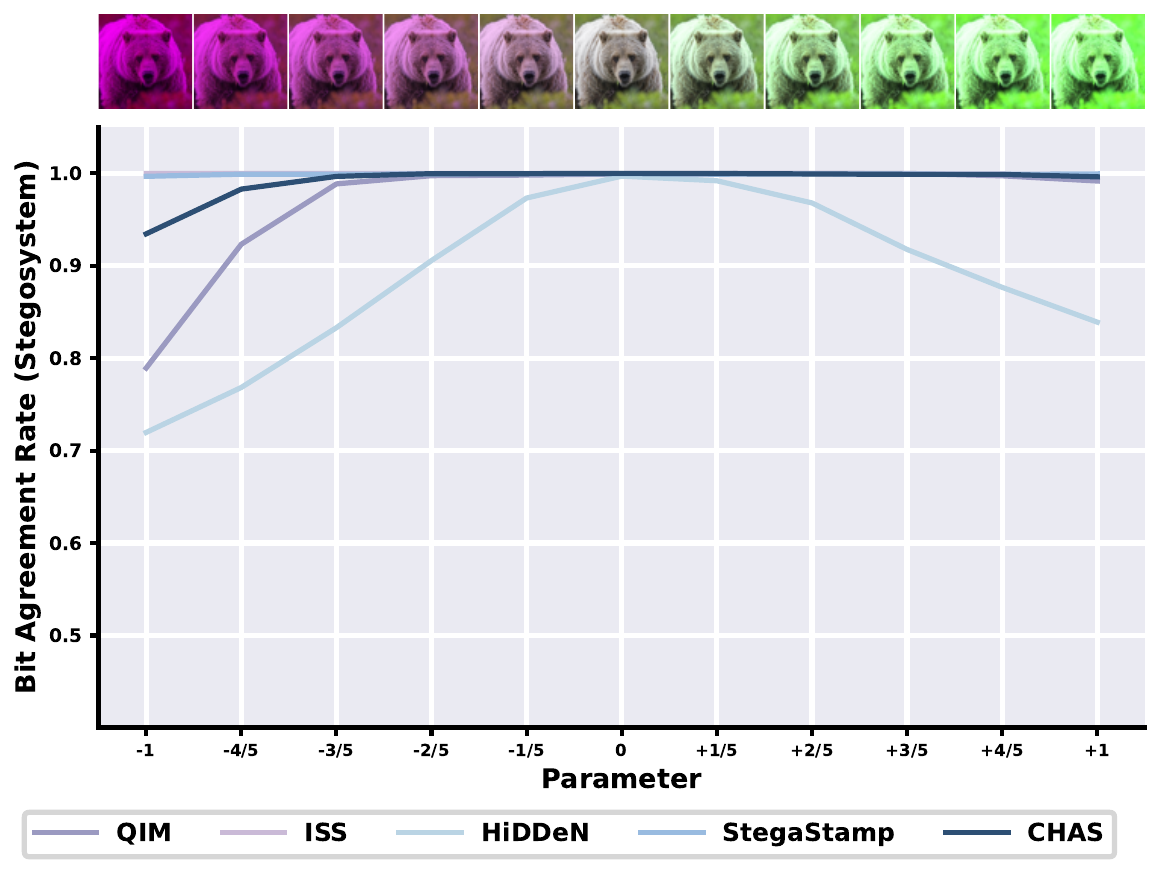}
    }
    \subfloat[Colour: Warmth]{
        \includegraphics[width=0.22\linewidth]{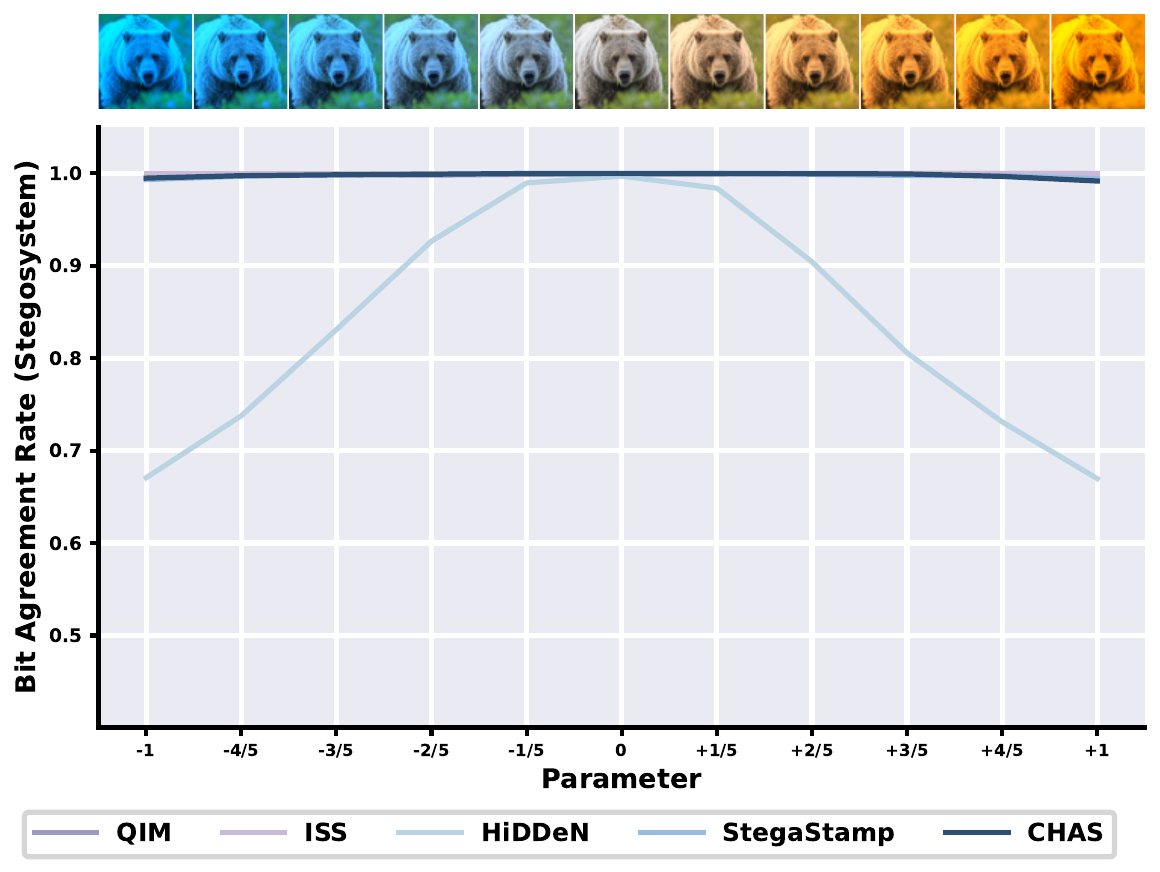}
    }
    \\
    \subfloat[Details: Blur]{
        \includegraphics[width=0.22\linewidth]{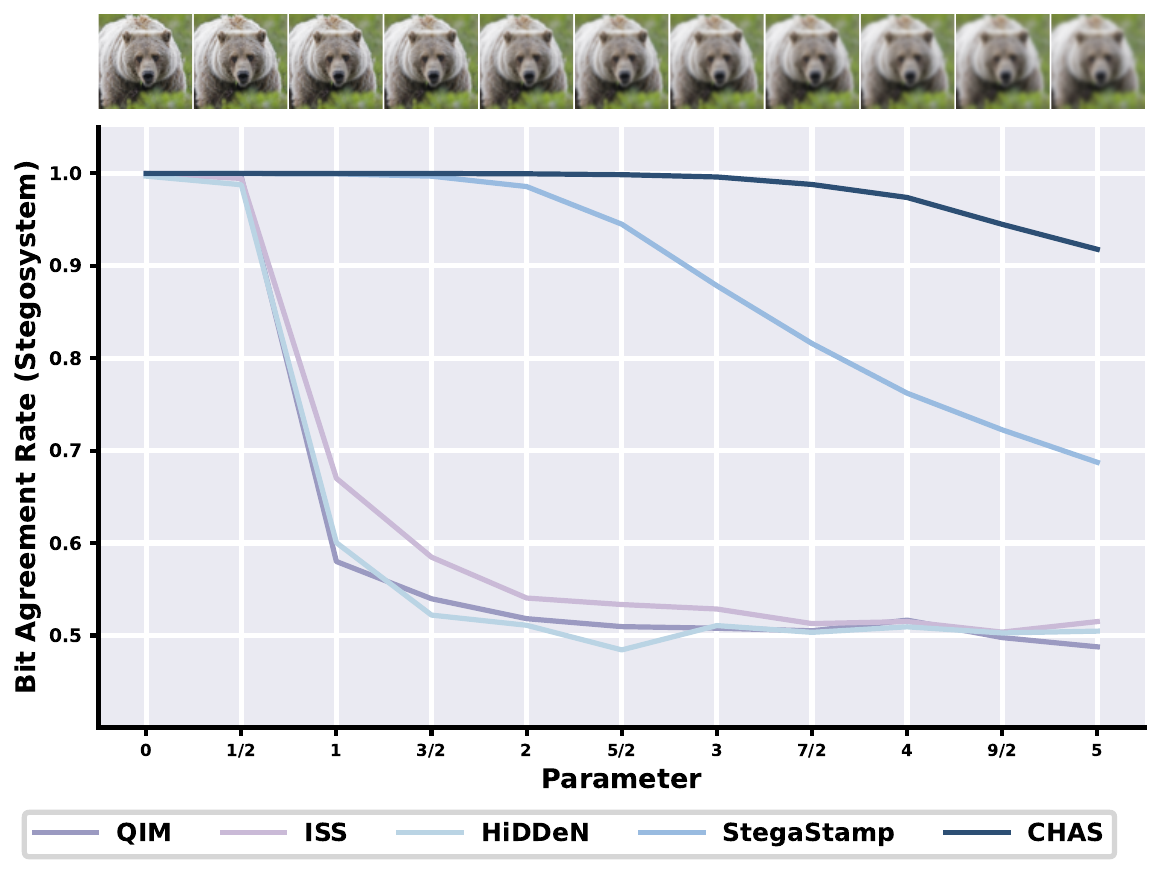}
    }
    \subfloat[Details: Grain]{
        \includegraphics[width=0.22\linewidth]{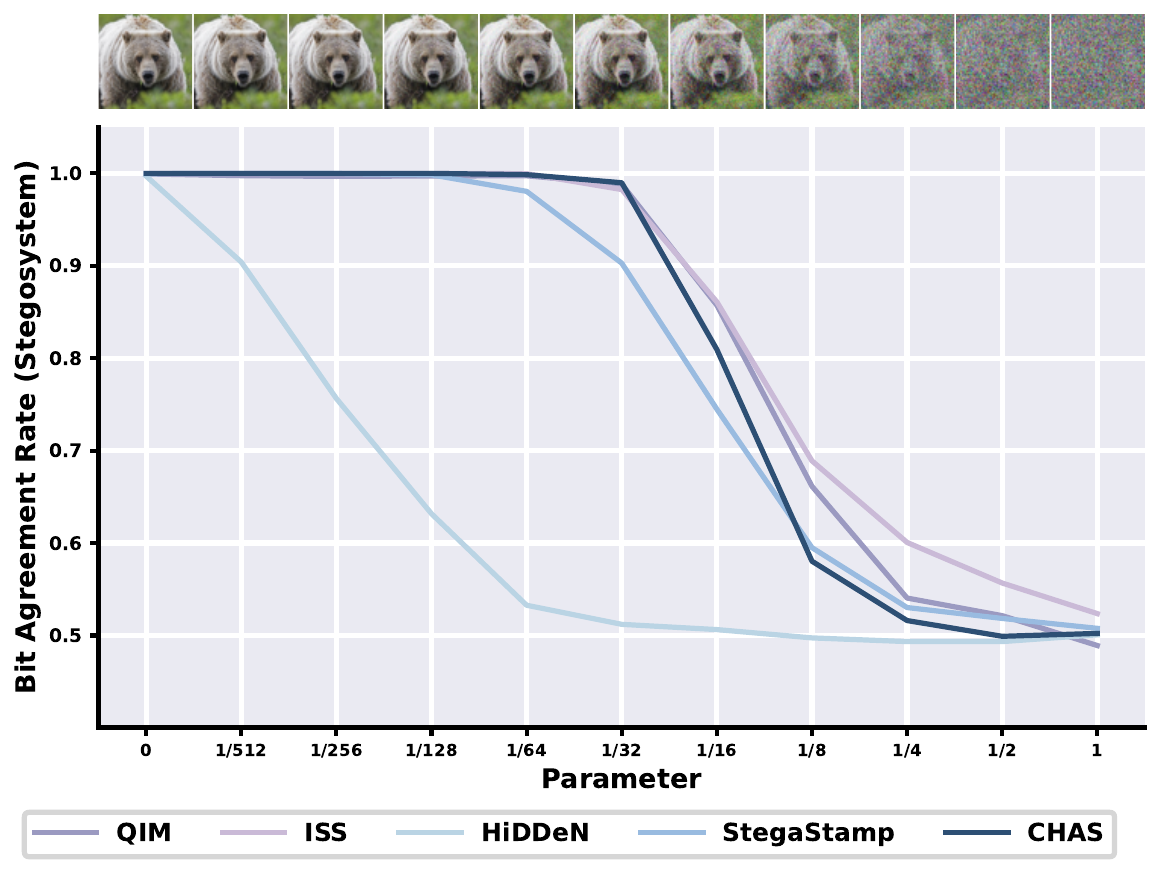}
    }
    \subfloat[Details: Sharpen]{
        \includegraphics[width=0.22\linewidth]{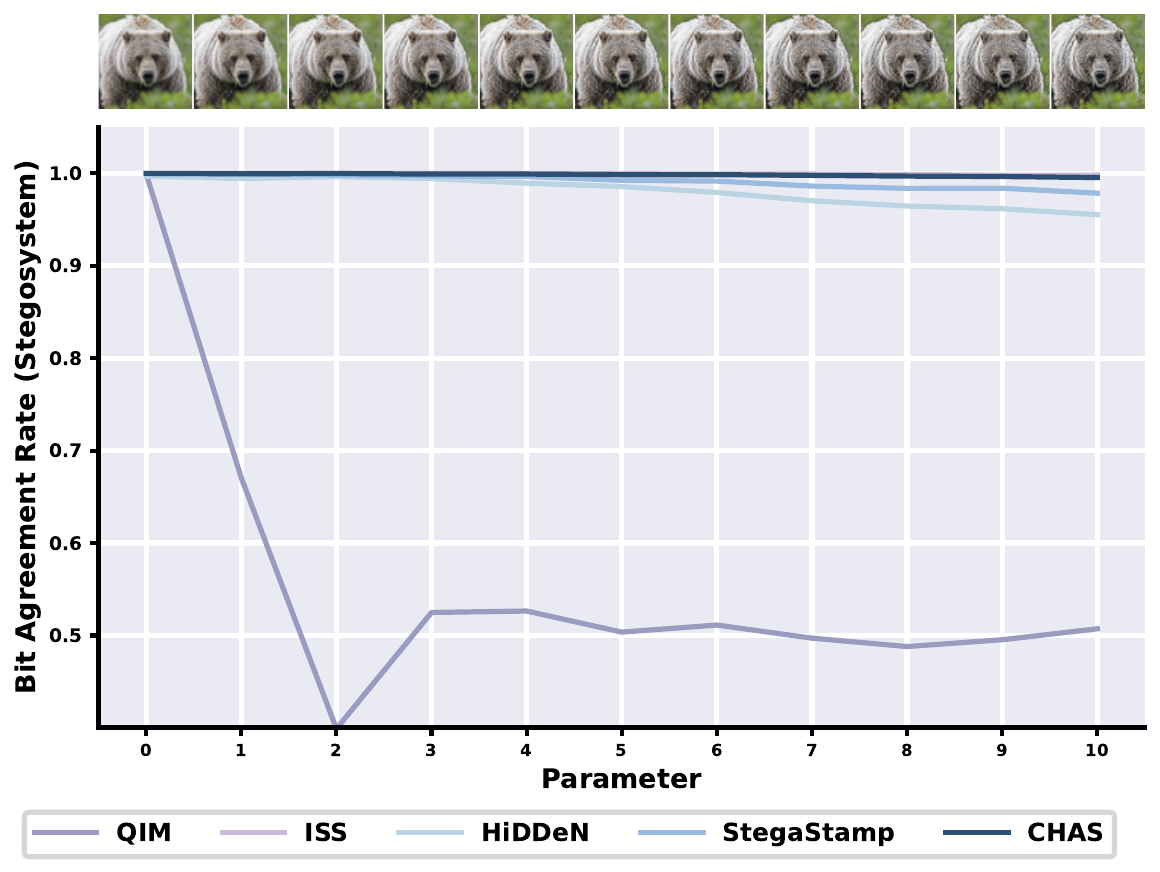}
    }
    \subfloat[Details: JPEG]{
        \includegraphics[width=0.22\linewidth]{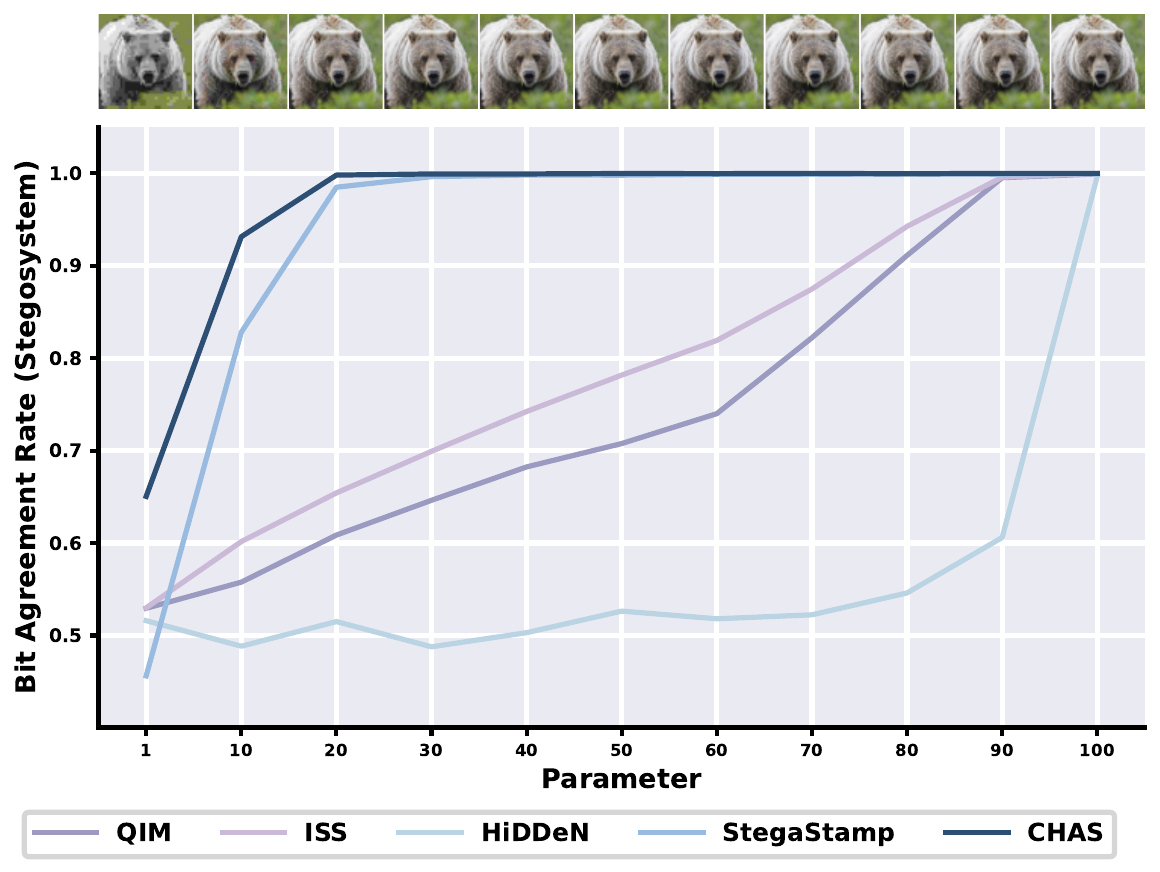}
    }
    \\
    \subfloat[Geometry: Crop]{
        \includegraphics[width=0.22\linewidth]{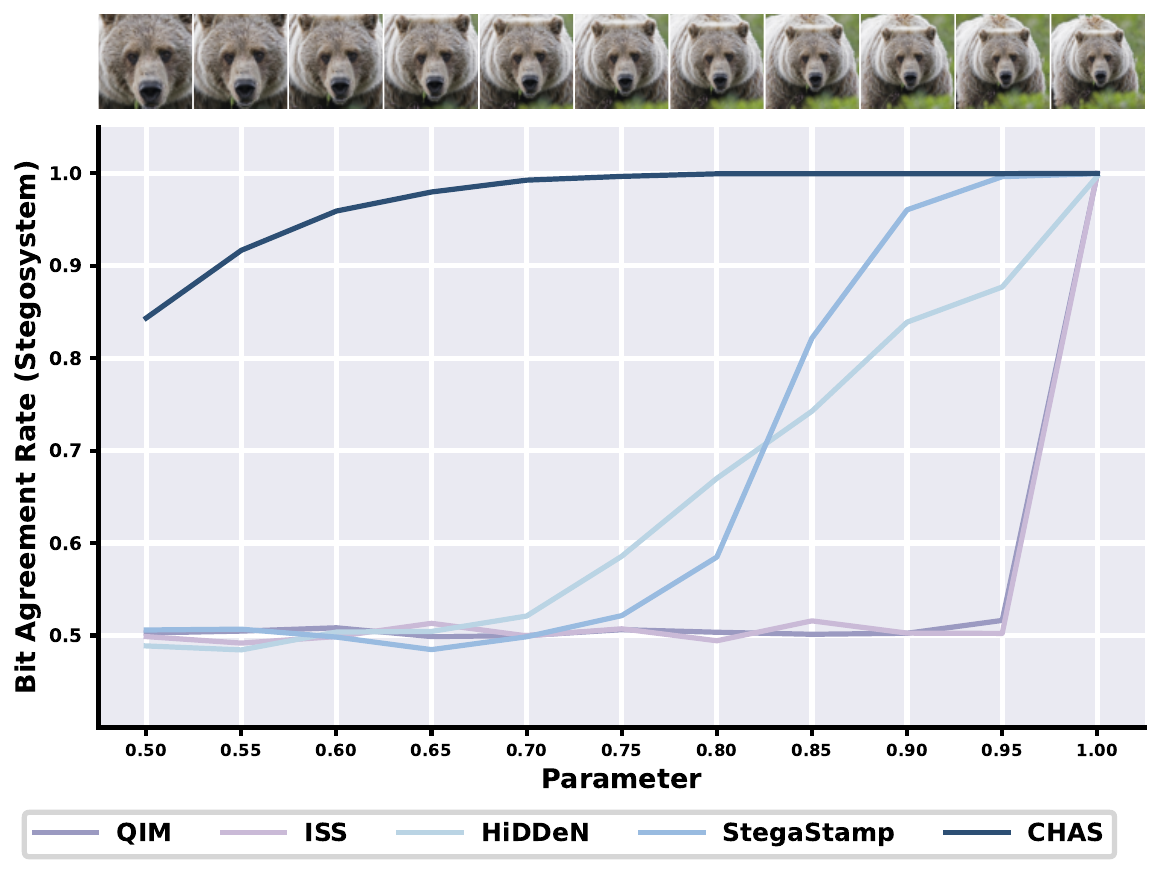}
    }
    \subfloat[Geometry: Rotate]{
        \includegraphics[width=0.22\linewidth]{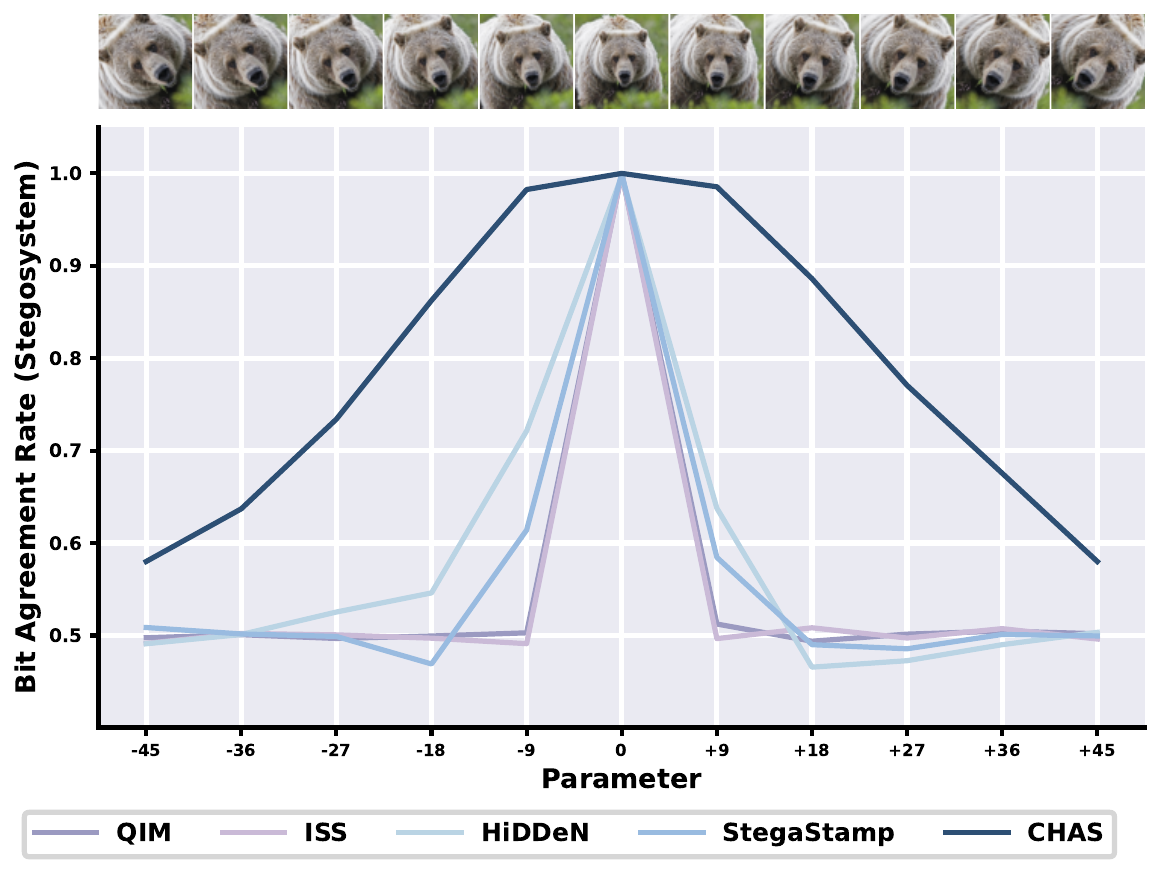}
    }
    \subfloat[Geometry: Horizontal Perspective]{
        \includegraphics[width=0.22\linewidth]{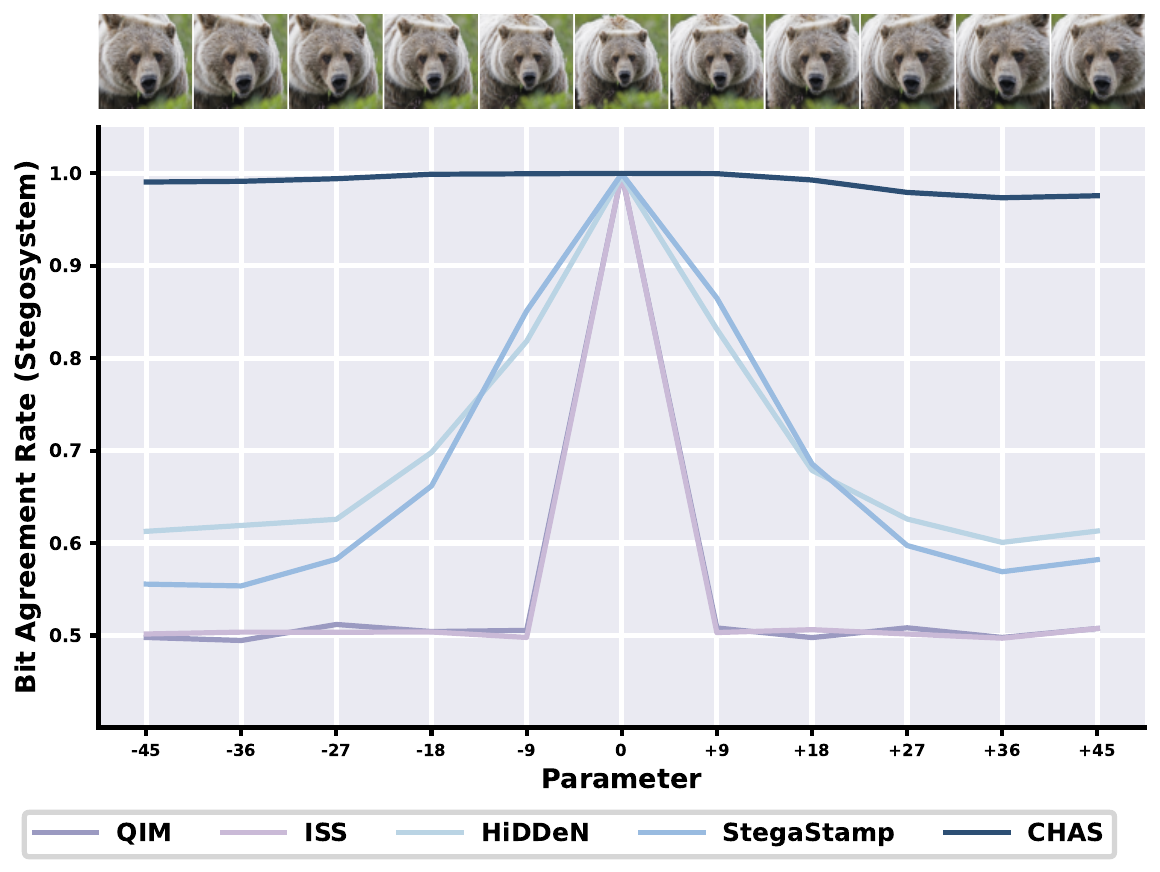}
    }
    \subfloat[Geometry: Vertical Perspective]{
        \includegraphics[width=0.22\linewidth]{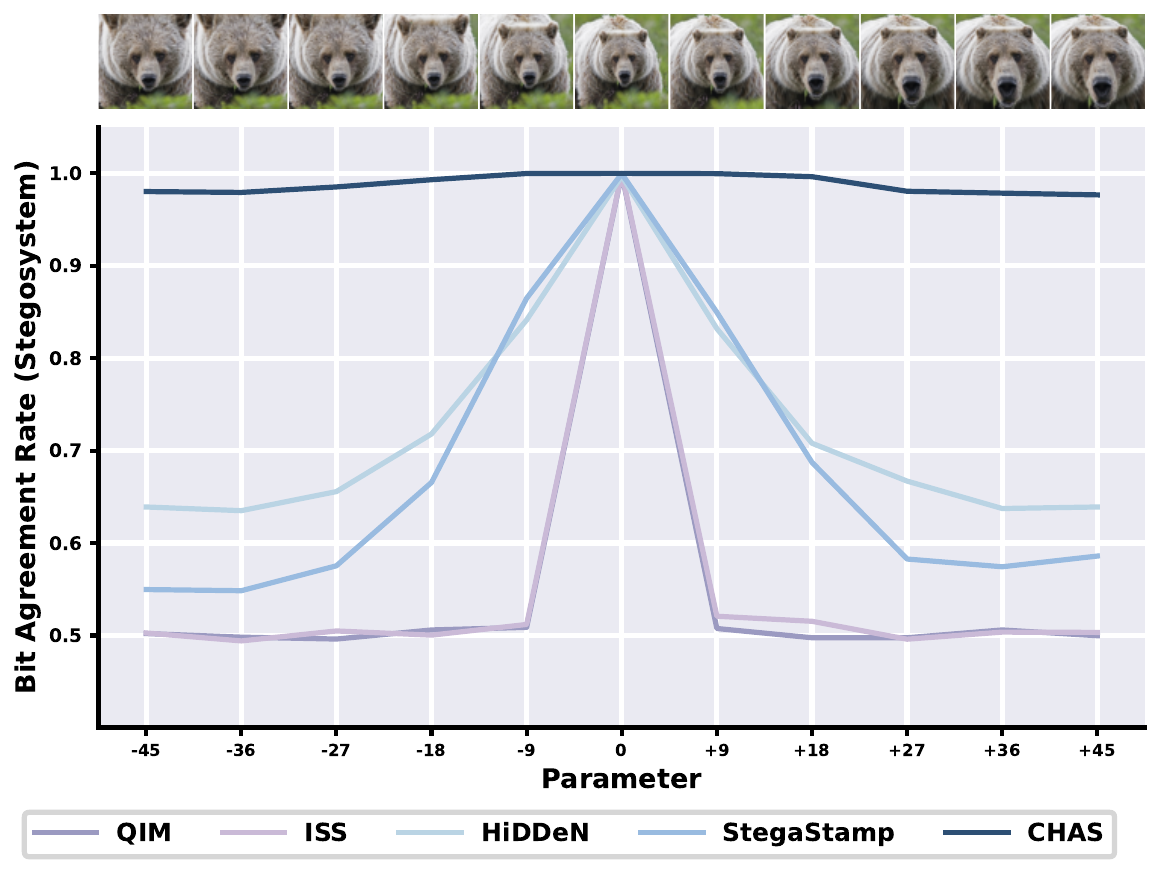}
    }
    \caption{Bit agreement rates of stegosystems across common processing operations.}
    \label{fig:stegosystem_common_edits}
\end{figure*}

\begin{figure}[!t]
\centering
\includegraphics[width=1.0\linewidth]{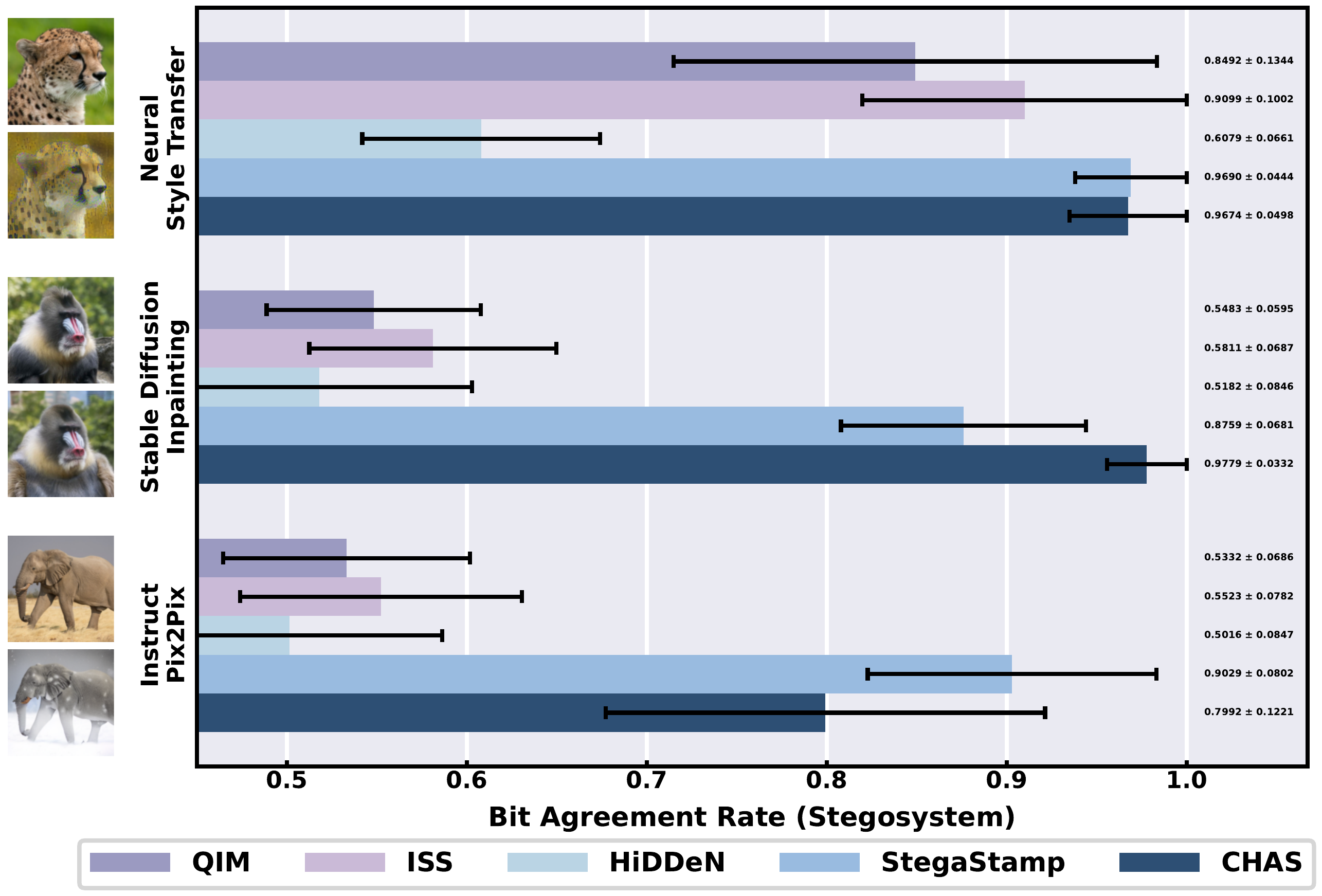}
\caption{Bit agreement rates of stegosystems under semantic editing.}
\label{fig:stegosystem_semantic_edits}
\end{figure}

\section{Evaluation}
The empirical evaluation that follows is organised around two questions. The first concerns the constituent parts, situating the proposed stegosystem amongst existing ones. The second concerns the system as a whole, namely the reliability with which the true parent-offspring pair is retrieved across the conditions that practical deployment may impose.

\begin{figure*}[t!]
    \centering
    \subfloat[Light: Brightness]{
        \includegraphics[width=0.23\linewidth]{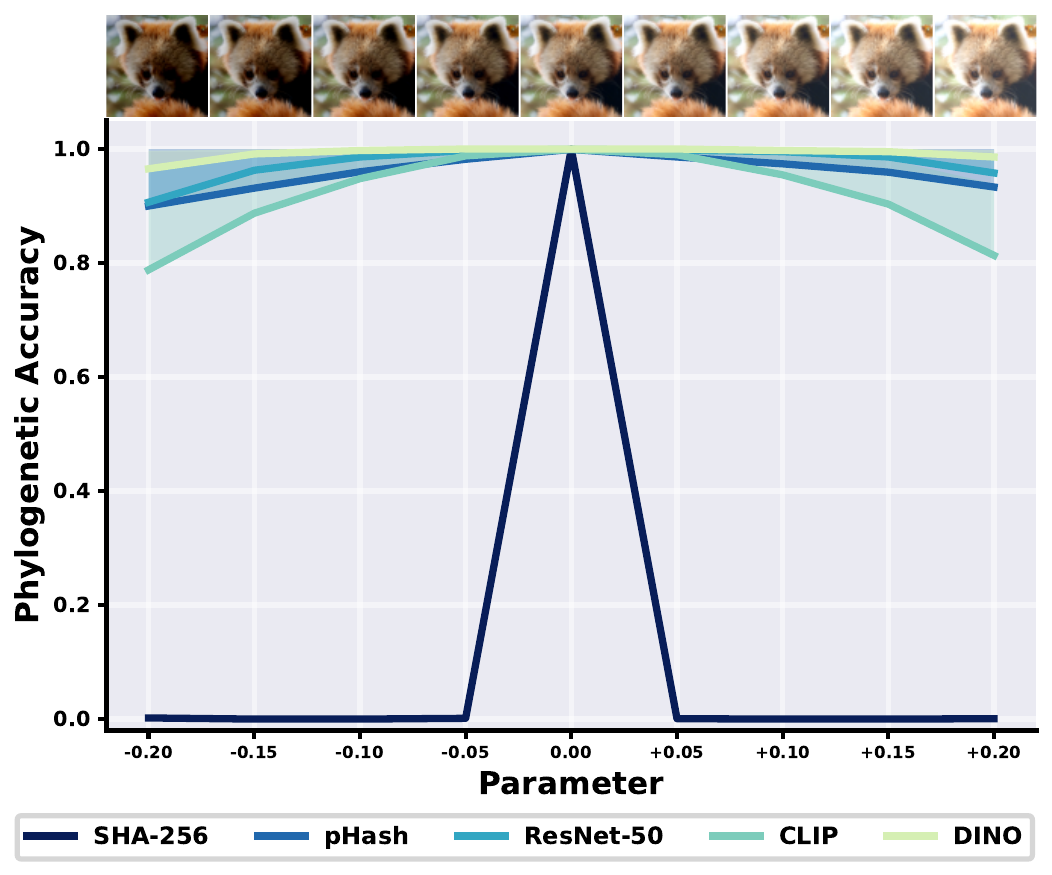}
    }
    \subfloat[Light: Contrast]{
        \includegraphics[width=0.23\linewidth]{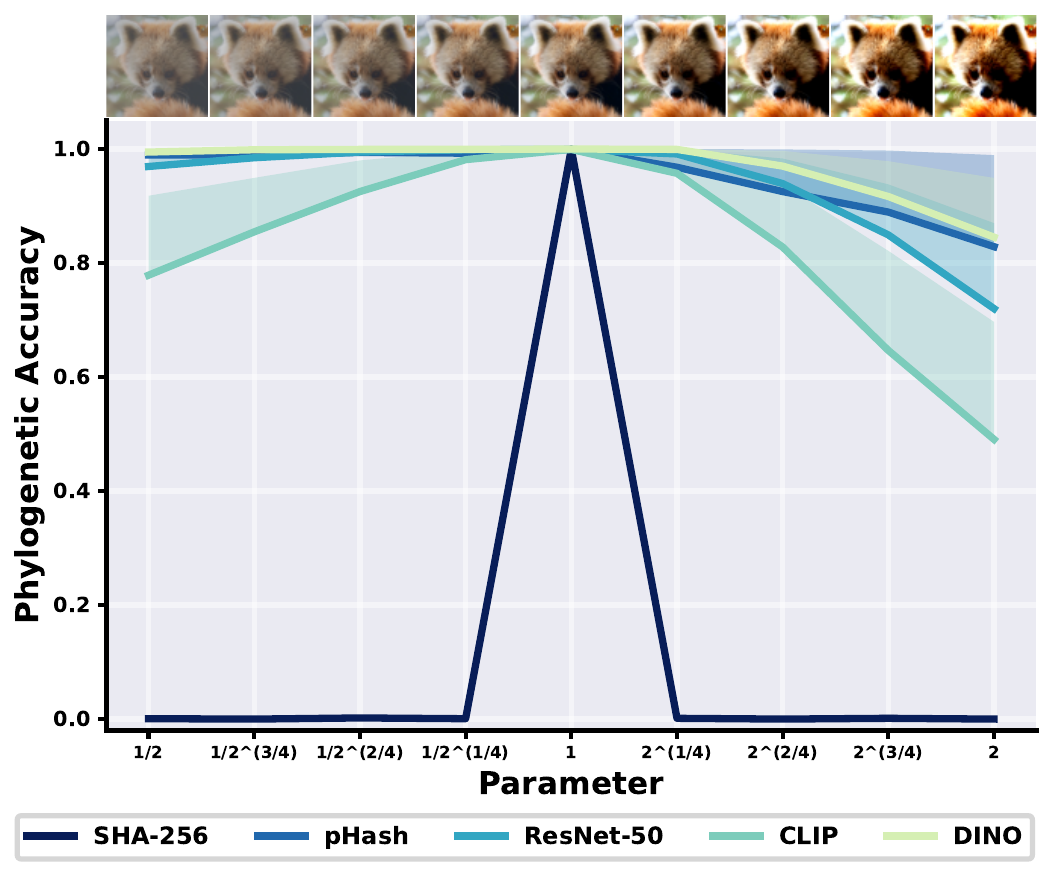}
    }
    \subfloat[Light: Exposure]{
        \includegraphics[width=0.23\linewidth]{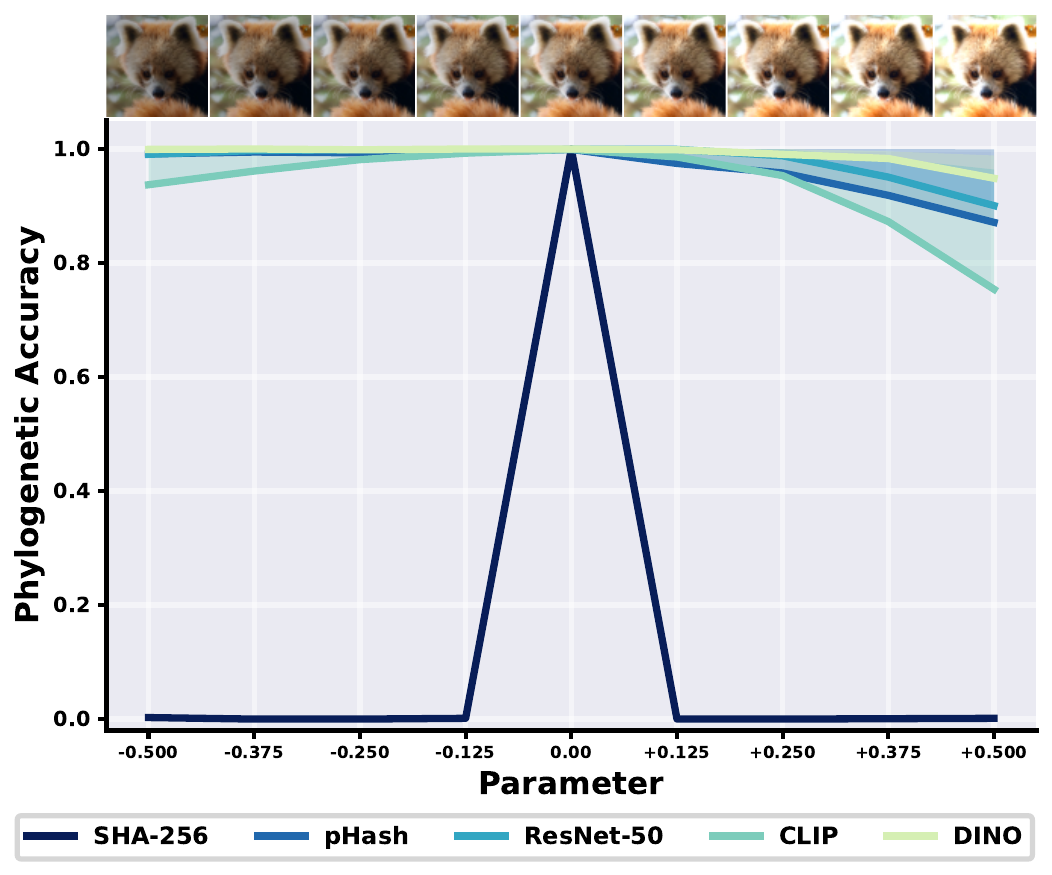}
    }
    \\
    \subfloat[Colour: Saturation]{
        \includegraphics[width=0.23\linewidth]{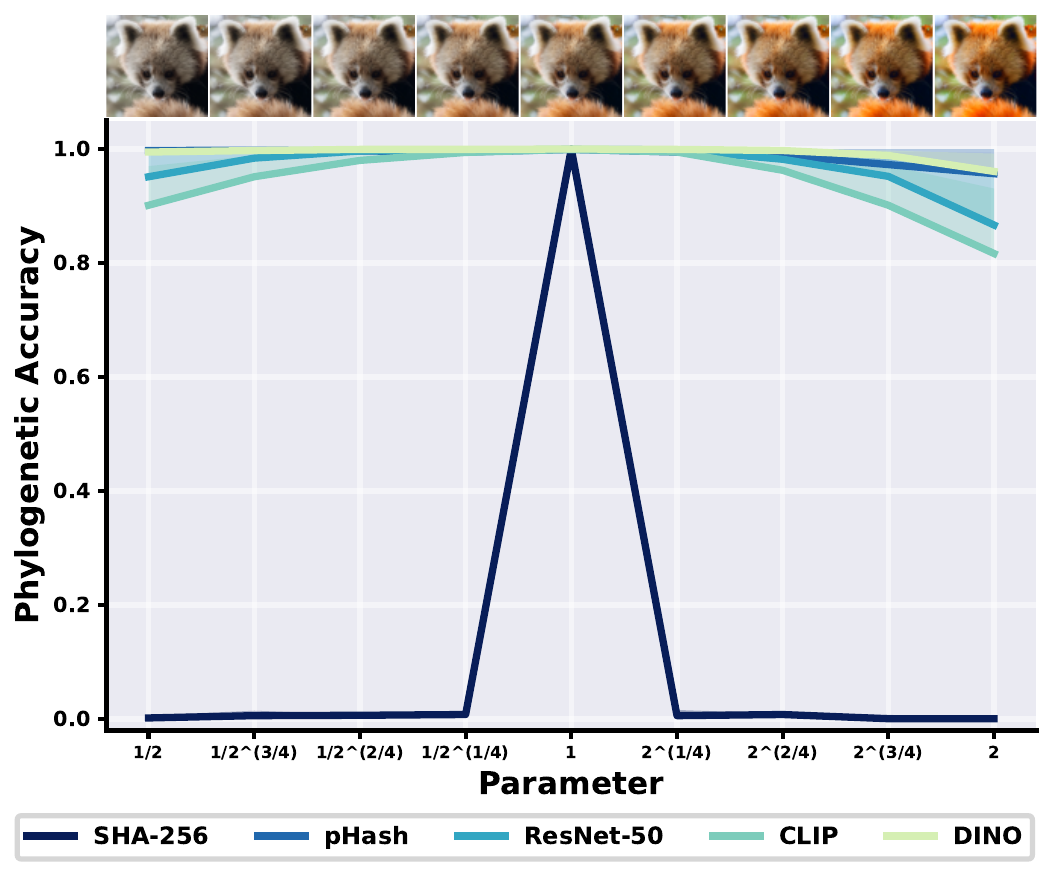}
    }
    \subfloat[Colour: Tint]{
        \includegraphics[width=0.23\linewidth]{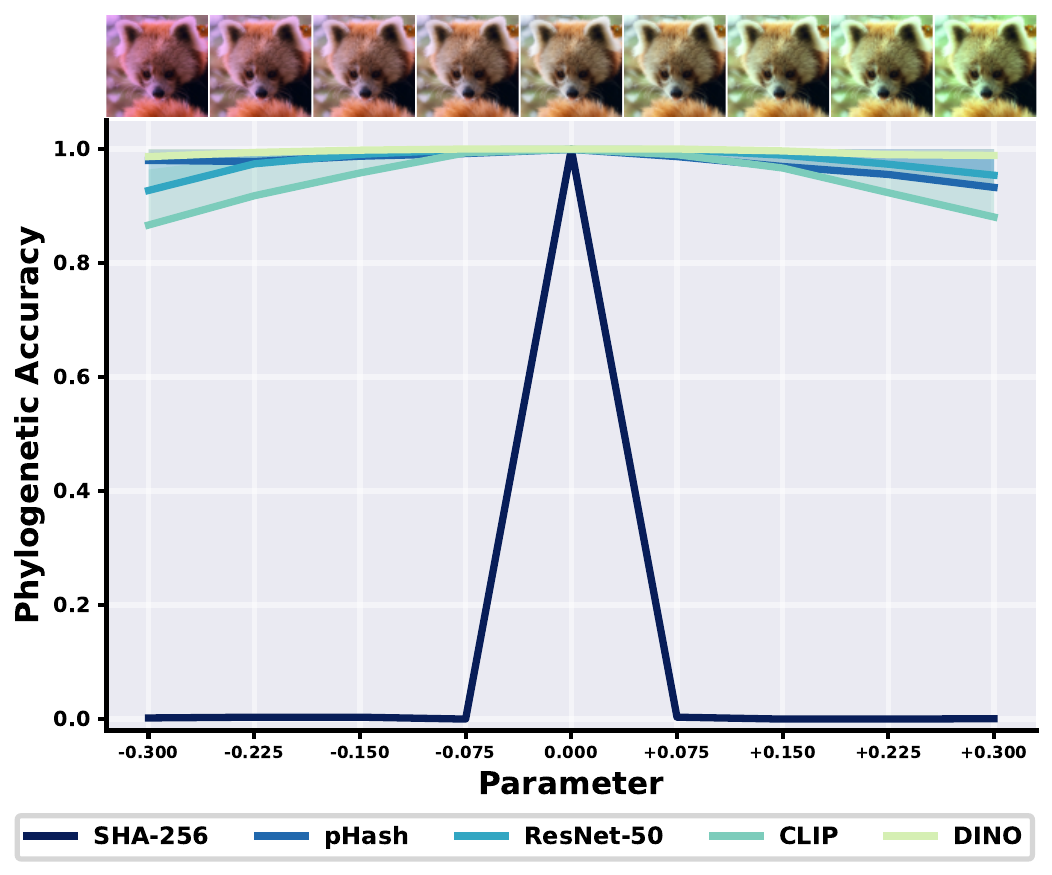}
    }
    \subfloat[Colour: Warmth]{
        \includegraphics[width=0.23\linewidth]{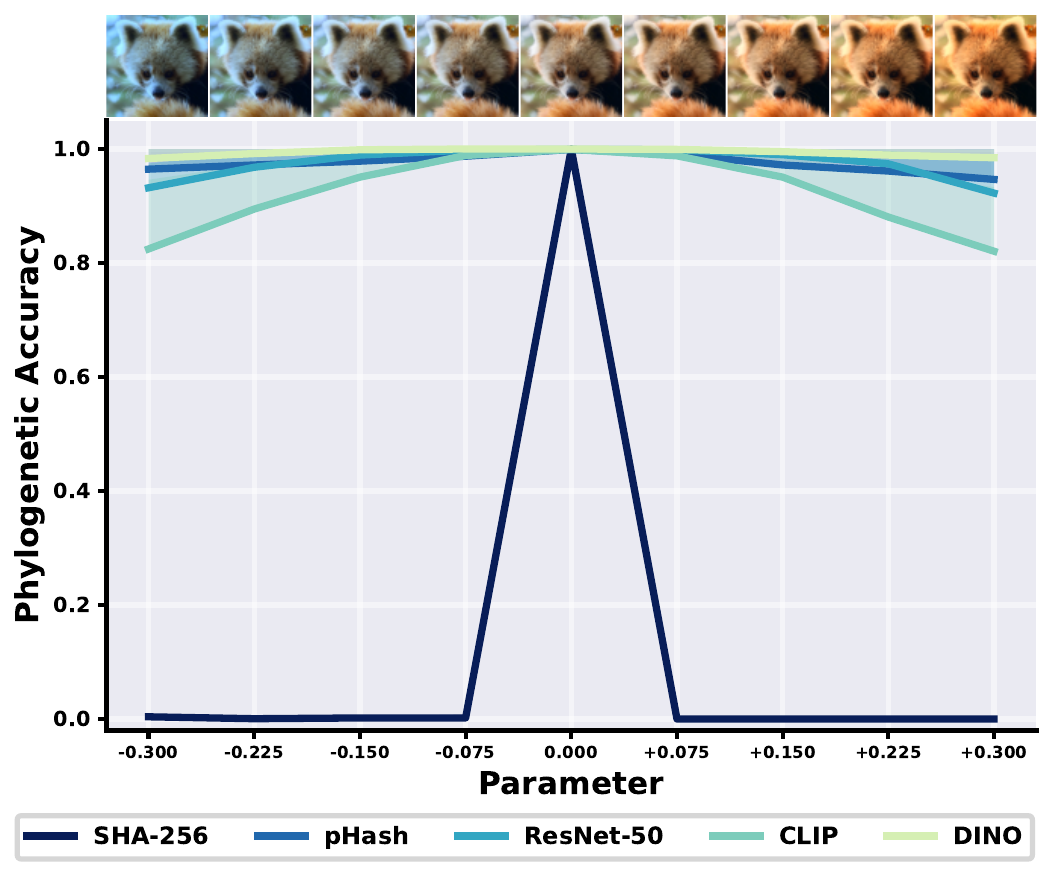}
    }
    \\
    \subfloat[Details: Blur]{
        \includegraphics[width=0.23\linewidth]{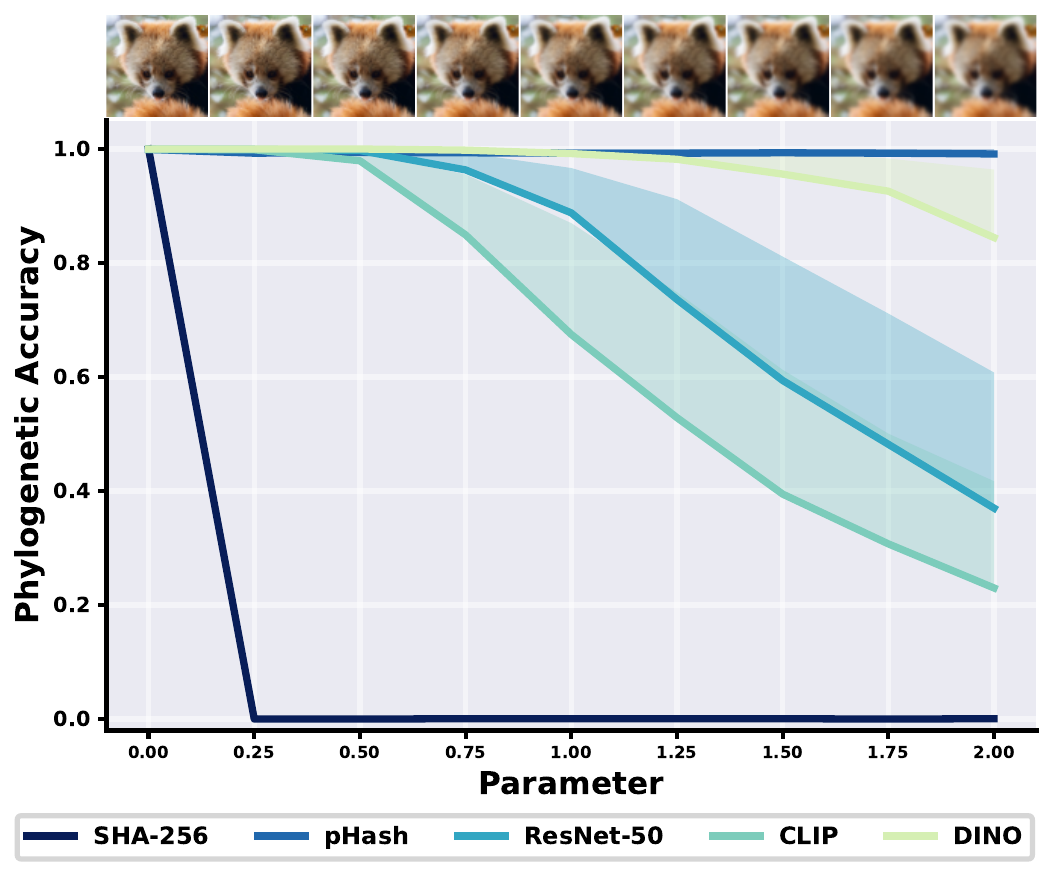}
    }
    \subfloat[Details: Grain]{
        \includegraphics[width=0.23\linewidth]{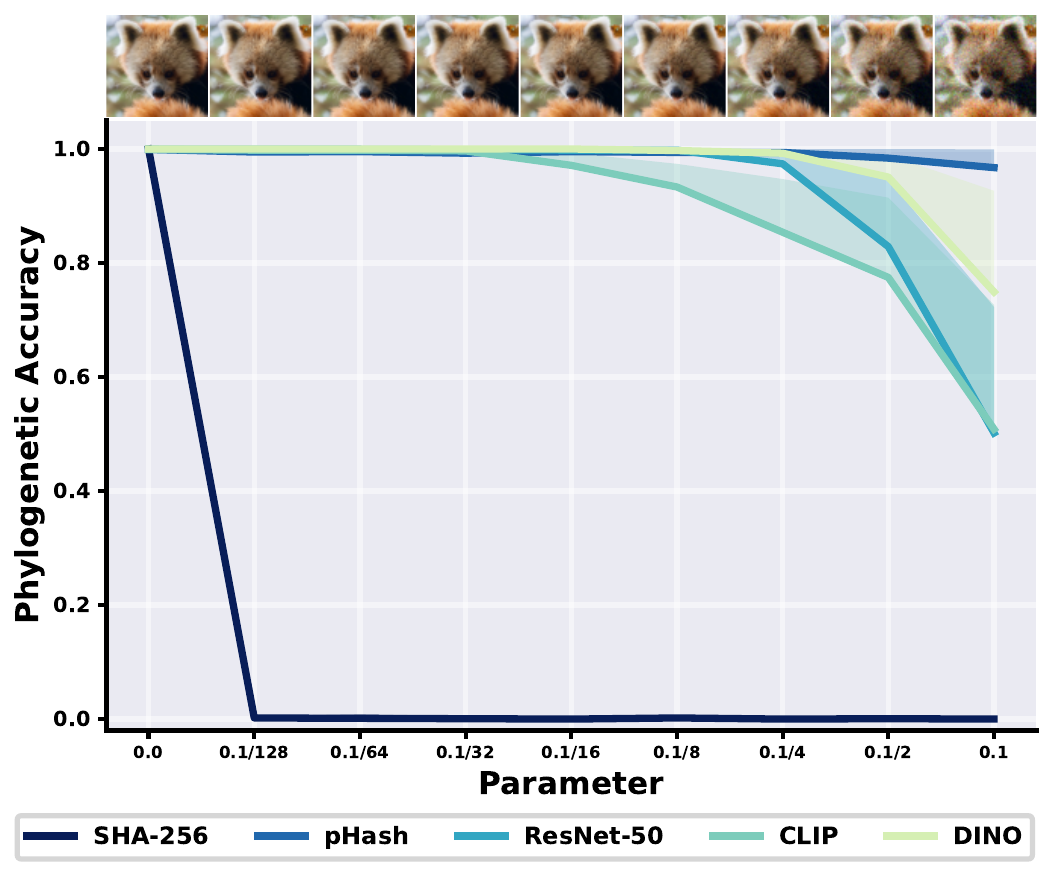}
    }
    \subfloat[Details: Sharpen]{
        \includegraphics[width=0.23\linewidth]{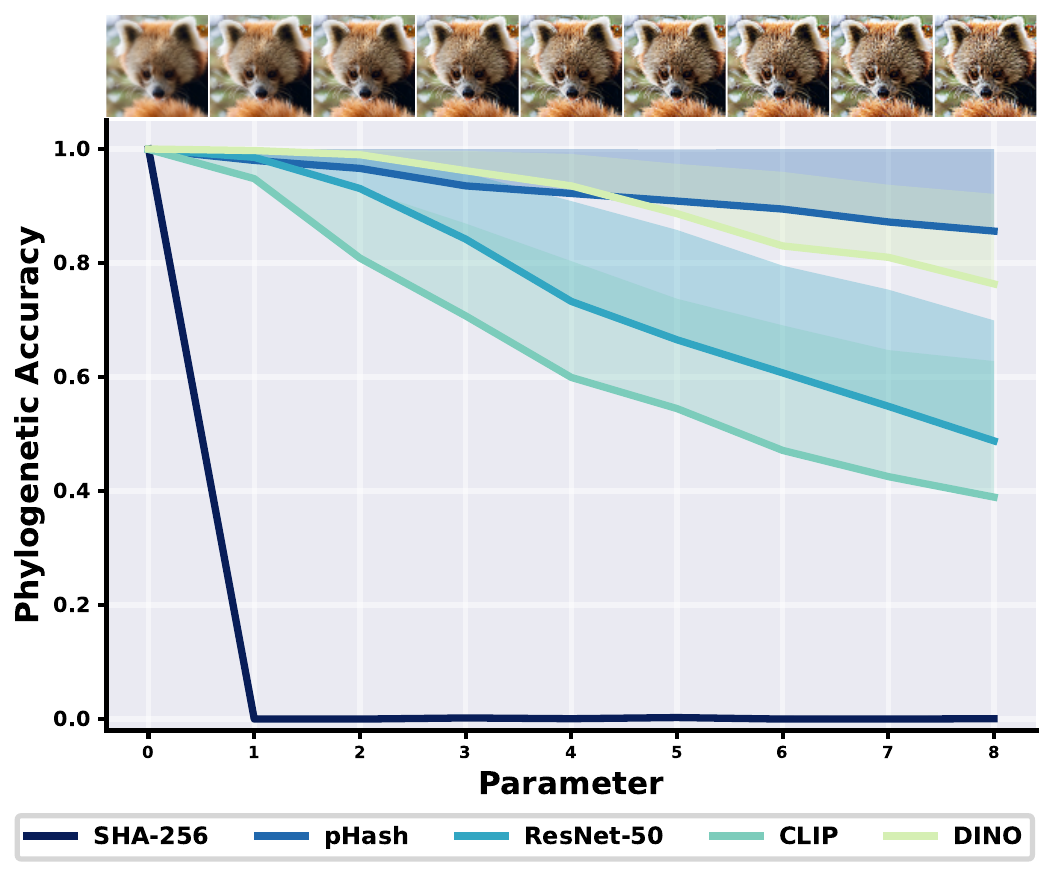}
    }
    \subfloat[Details: JPEG]{
        \includegraphics[width=0.23\linewidth]{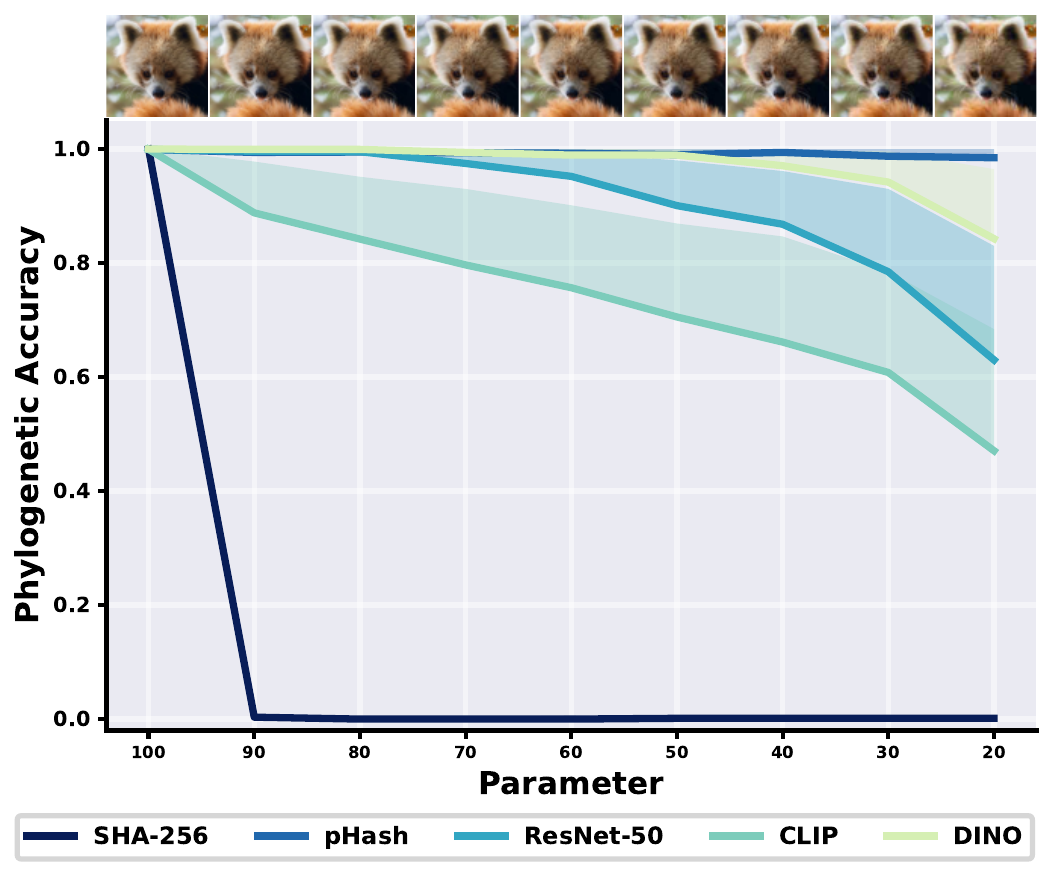}
    }
    \\
    \subfloat[Geometry: Crop]{
        \includegraphics[width=0.23\linewidth]{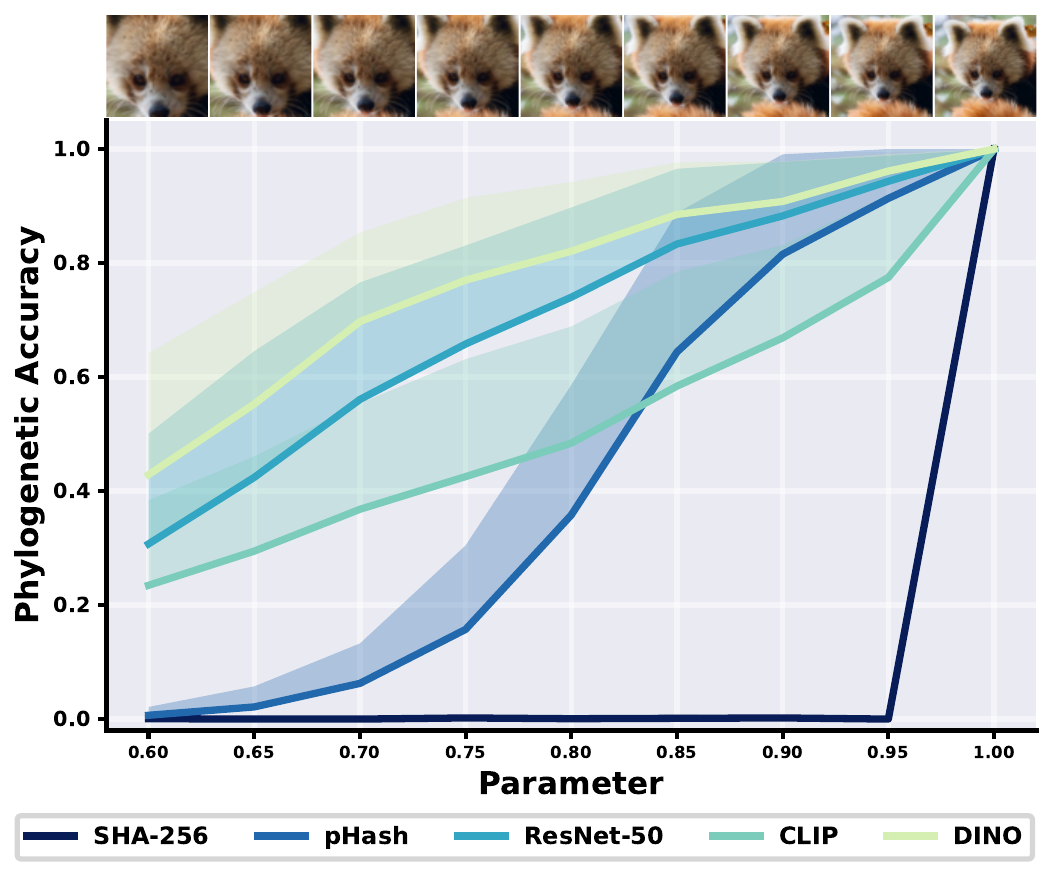}
    }
    \subfloat[Geometry: Rotate]{
        \includegraphics[width=0.23\linewidth]{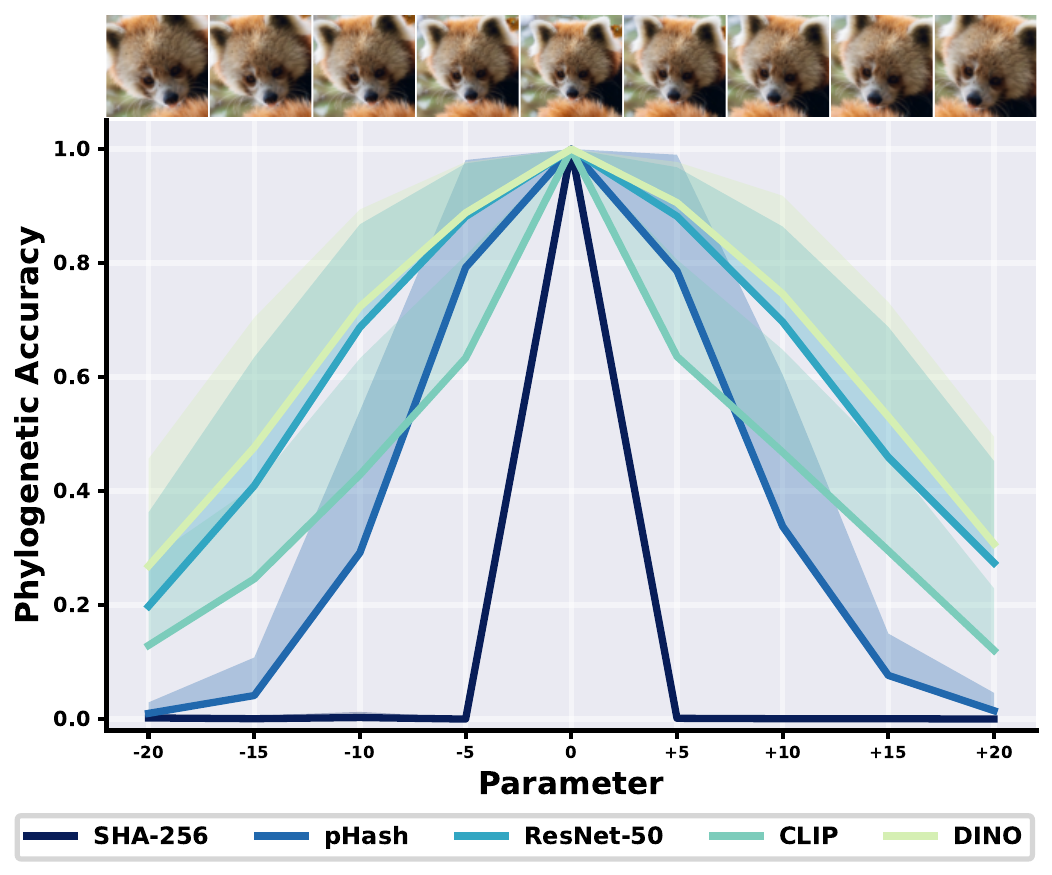}
    }
    \subfloat[Geometry: Horizontal Perspective]{
        \includegraphics[width=0.23\linewidth]{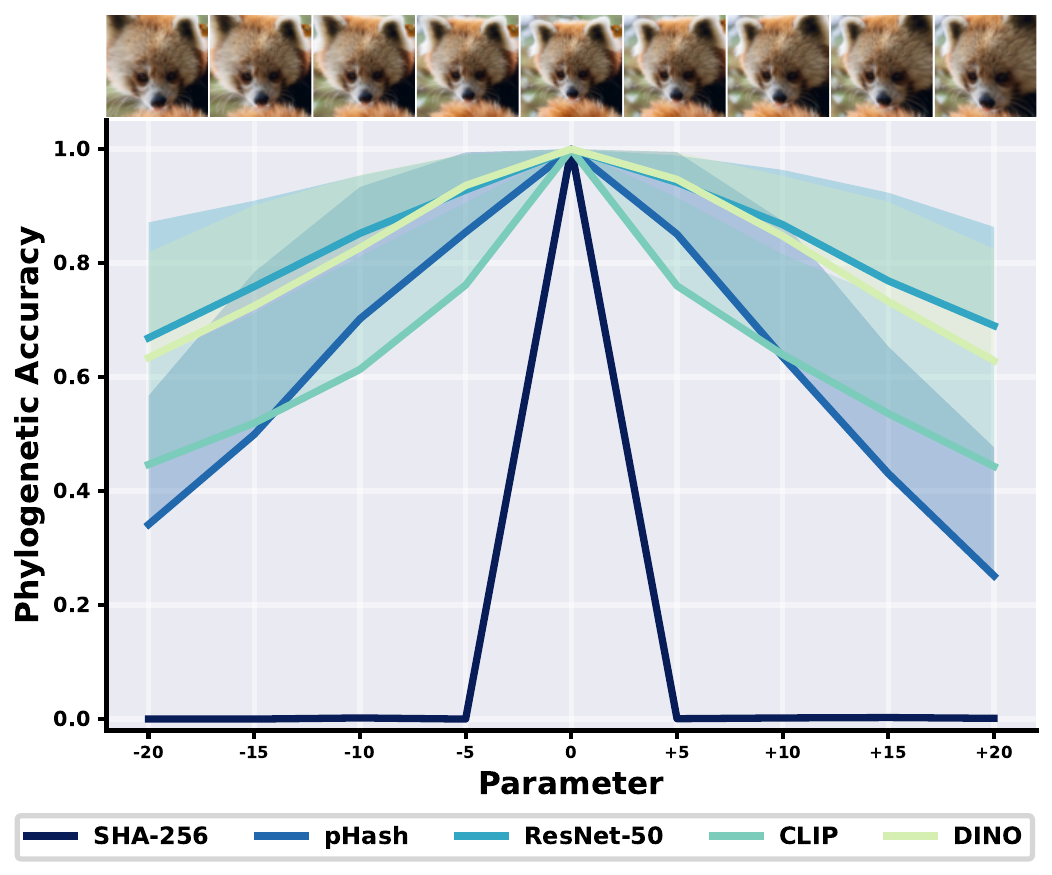}
    }
    \subfloat[Geometry: Vertical Perspective]{
        \includegraphics[width=0.23\linewidth]{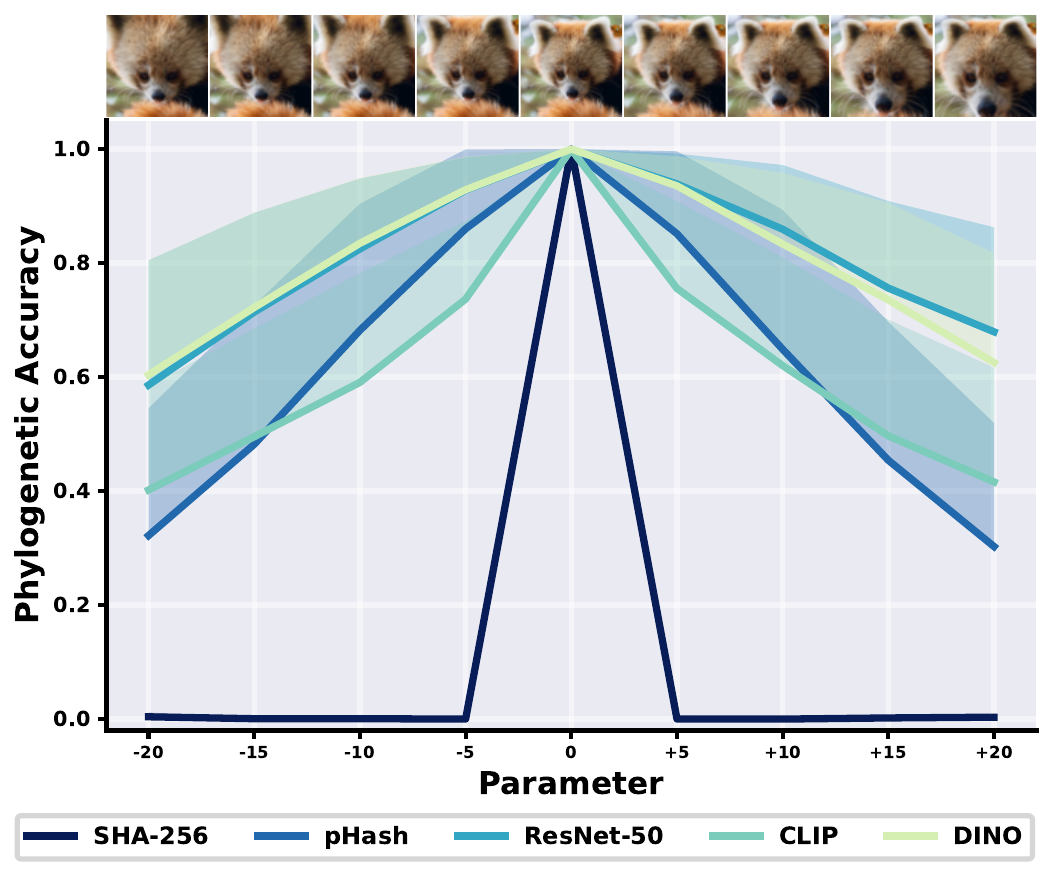}
    }
    \caption{Phylogenetic accuracy across common processing operations.}
    \label{fig:phylogeny_common_edits}
\end{figure*}

\begin{figure*}[t!]
    \centering
    \subfloat[Heatmaps (Precision, Recall and F-score)]{
        \includegraphics[width=0.98\linewidth]{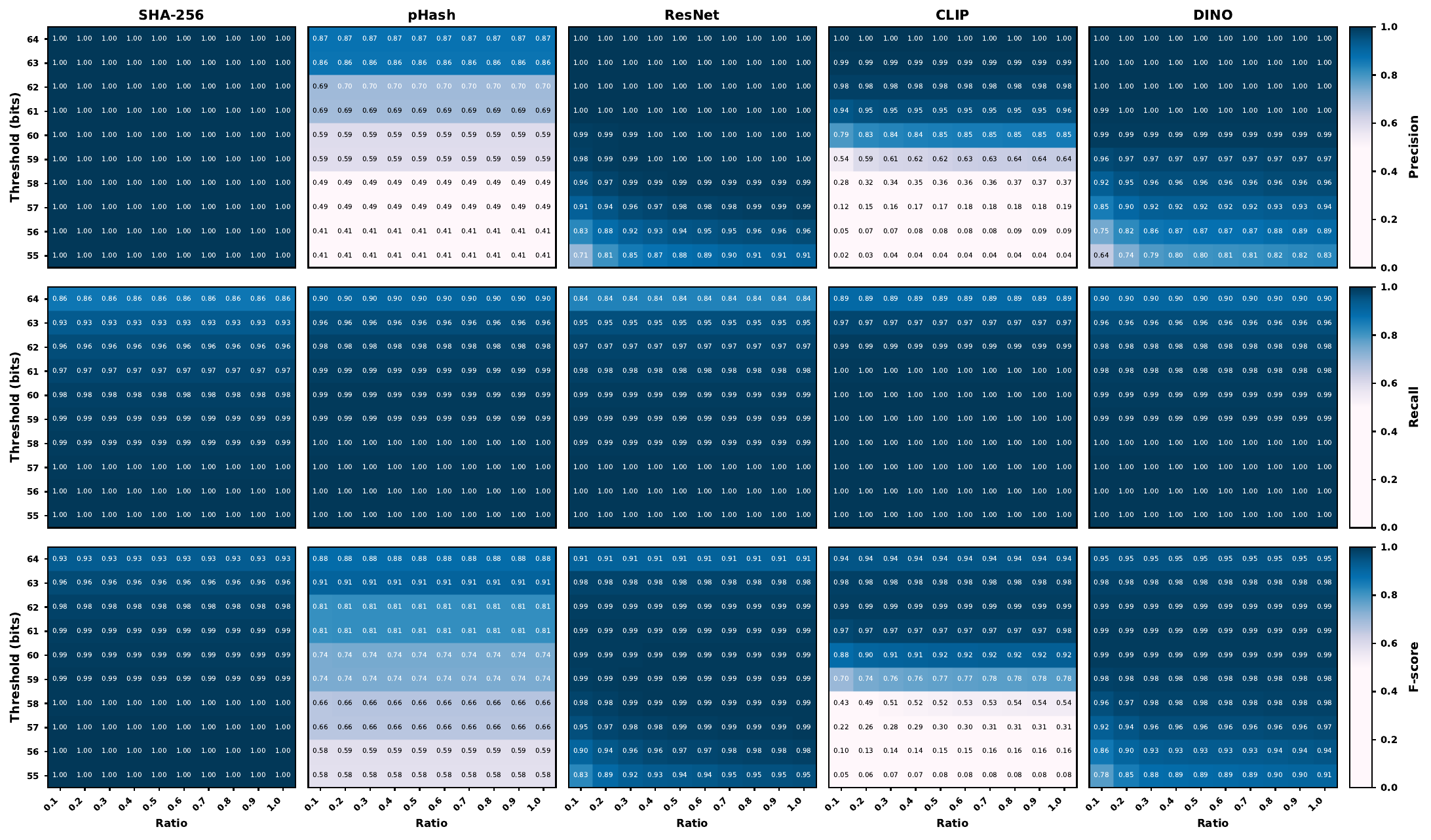}
    }
    \\
    \subfloat[F-Score Benchmark]{
        \includegraphics[width=0.98\linewidth]{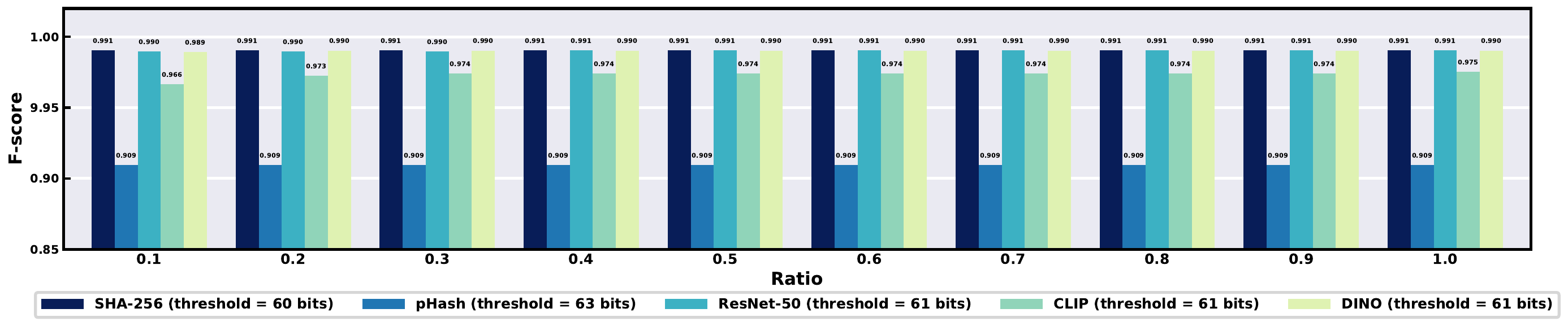}
    }
    \caption{Precision, recall and F-score of phylogenetic retrieval under inclusion of extraneous samples.}
    \label{fig:inclusion}
\end{figure*}

\begin{figure*}[t!]
    \centering
    \subfloat[Heatmaps (Precision, Recall and F-score)]{
        \includegraphics[width=0.98\linewidth]{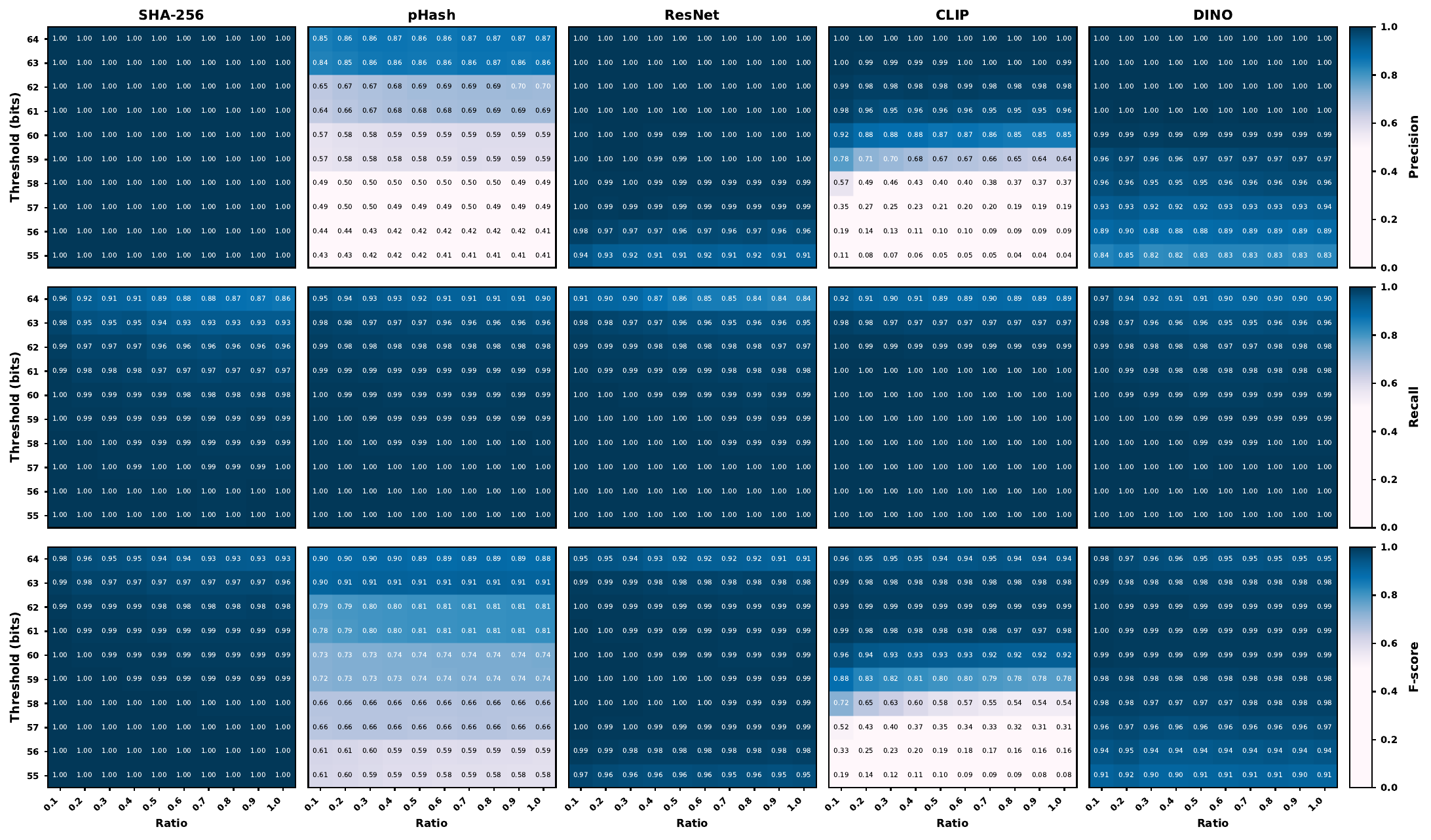}
    }
    \\
    \subfloat[F-Score Benchmark]{
        \includegraphics[width=0.98\linewidth]{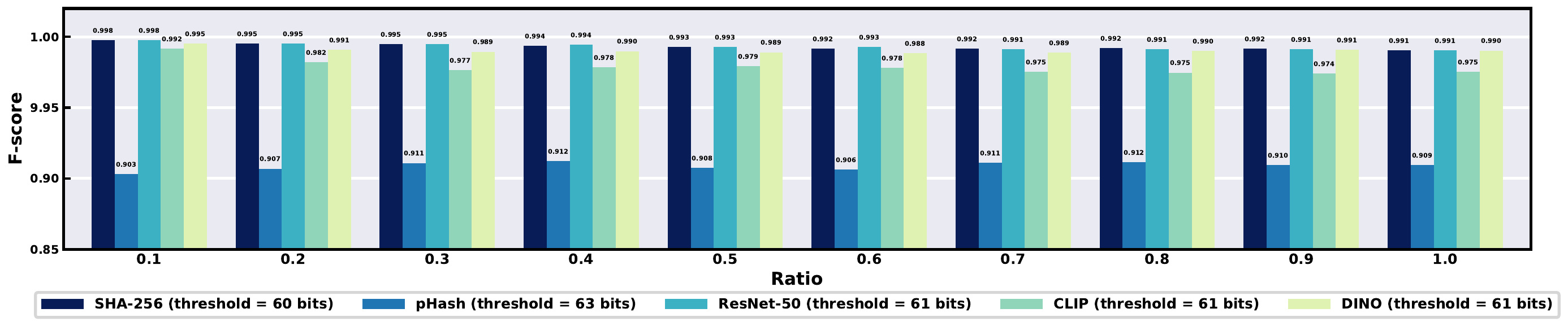}
    }
    \caption{Precision, recall and F-score of phylogenetic retrieval under deletion of relevant samples.}
    \label{fig:deletion}
\end{figure*}

\subsection{Analysis of Stegosystems}
We begin by examining the stegosystems in isolation, setting aside for the moment the choice of projector. The concern here is a more elementary one: how faithfully does each system embed and extract a trait under various conditions, and at what cost to the fidelity of the content it inhabits?

Table~\ref{tab:stegosystem_performance} reports the performance of each stegosystem under conditions where stego images remain unmodified. All methods attain near-perfect bit accuracy, though the distortion introduced to the content varies among them. The classical methods, QIM and ISS, introduce distortions that are, by all four image quality measures, among the least perceptible. Among the learning-based methods, CHAS and HiDDeN introduce comparable levels of distortion, with StegaStamp incurring the most visible degradation to image quality.

Table~\ref{tab:stegosystem_complexity} reports the model size of each stegosystem. The parameter count of the classical methods is nearly negligible, as their operation involves little more than a pseudo-random seed for determining the carrier coefficients, the associated projection vectors and other auxiliary information. The learning-based methods, by contrast, require their full complement of neural network weights for deployment. Among these, HiDDeN is the most compact, CHAS falls in between, and StegaStamp demands the most memory, owing in large part to the spatial transformer network it employs.

We turn next to how robust each stegosystem is against common processing operations. Figure~\ref{fig:stegosystem_common_edits} reports the bit agreement rate of each stegosystem across a broad spectrum of operations, as the severity of each adjustment is varied from its neutral value outward. The classical methods, by virtue of operating in the DCT domain of the luminance channel, are naturally insensitive to colour adjustments and comparably resilient to moderate lighting changes. The primary limitation of classical methods lies in their susceptibility to geometric transformations. Conversely, learning-based methods, having been trained to handle such distortions, demonstrate a clear advantage. Among them, the proposed CHAS achieves the most sustained robustness across the full range of geometric operations, whilst maintaining competitive performance under adjustments to lighting, colour and details.

We turn finally to the case of semantic editing, wherein the content undergoes transformation not merely in its photometric or geometric properties, but in its underlying semantic meaning. Such transformations arise when media, once released, undergoes generative processes outside the cooperative framework, yet against which the system aspires to remain resilient. Figure~\ref{fig:stegosystem_semantic_edits} reports the bit agreement rate of each stegosystem under three modes of semantic editing. The classical methods yield a surprising result in that, even after style transfer, these early techniques manage to recover the payload well above chance. For local editing with inpainting, the proposed CHAS achieves the highest bit agreement rate, followed by StegaStamp, while the remaining methods fall near chance. For global edits of a seasonal or atmospheric nature, StegaStamp achieves the highest bit agreement rate, whilst CHAS proves less adept in this regard, and the remaining methods fall near random guessing. This outcome may be attributed to StegaStamp's intended deployment in physical-world environments, where global variations in atmospheric conditions, such as snowfall, rainfall and fog, are commonplace, and its training might, implicitly or explicitly, reflect this expectation.

It is worth pausing to reflect on what these results, taken together, suggest. The classical methods, conceived decades ago, grounded upon principles of signal processing, exhibit a robustness to colour and lighting adjustments, and even to certain semantic transformations by generative models, that is far from negligible. This serves as a reminder that the efficacy of a design is not dictated by its recency. The learning-based methods, meanwhile, demonstrate a more consistent robustness to geometric transformations and point towards the potential for resisting semantic edits powered by generative AI.

\subsection{Analysis of Phylogenetic System}
For the phylogenetic system as a whole, the concern is how reliably the true parent-offspring pairs may be retrieved under conditions that depart from the ideal: when the content has been edited, when the pool has been diluted with unrelated samples, or when parts of it are no longer present.

Figure~\ref{fig:phylogeny_common_edits} reports the phylogenetic accuracy of each projector, paired with the proposed stegosystem, across a broad range of common processing operations. In this experiment, all parent and offspring samples in the pool are edited at a given severity, and retrieval is performed without abstention, nominating the closest candidate from the pool. When no distortion is present, all projectors attain near-perfect phylogenetic accuracy, making the choice of projector inconsequential. As the severity of each operation increases, however, the projectors diverge markedly. By design, the sensitivity of SHA-256 to any perturbation is absolute, causing it to collapse under even mild distortion. While this property serves cryptographic integrity well, it fails to tolerate the variations found in a wild cyber ecosystem. Among the remaining projectors, robustness varies by operation type. All projectors remain largely unaffected by light and colour adjustments. For operations on details, DINO and pHash demonstrate greater resilience than the rest, while for operations on geometry, DINO and ResNet do so.

In practice, the pool assembled at the time of enquiry is drawn from a broader cyber ecosystem, where samples generated within the cooperative platform coexist with those of unrelated origin. Consequently, the proportion of relevant samples within such a pool may be small and not all of them may remain accessible. Figures~\ref{fig:inclusion} and~\ref{fig:deletion} examine the phylogenetic system's behaviour when the pool under inspection departs from the standard configuration, either by the introduction of extraneous samples, or by the removal of members from it. In both the inclusion and deletion experiments, retrieval is framed as a binary classification task. For each candidate-query pair, the system accepts the candidate as the parent if and only if the similarity score meets or exceeds a predefined threshold, and rejects it otherwise. Precision is the fraction of true pairs amongst all pairs claimed by the system. Recall is the fraction of true pairs successfully identified by the system amongst all true pairs in the pool.

In the inclusion experiment, the ratio denotes the proportion of the relevant samples amongst all. As the ratio decreases, more extraneous samples are added to the pool. These extraneous samples, though unrelated, may produce similarity scores that exceed the threshold by chance, increasing the number of pairs claimed by the system whilst the number of true pairs remains fixed. Precision therefore declines with the ratio, whilst recall is unaffected.

In the deletion experiment, the ratio denotes the fraction of relevant samples retained. As the ratio decreases, samples are progressively removed, reducing the number of visually related candidates from a common lineage. This can diminish the likelihood of false claims, as fewer confusable pairs remain to exceed the threshold. Therefore, precision may tend to rise as the ratio decreases, although this is not guaranteed. Likewise, recall is not guaranteed to follow a particular trend. Removing a true pair that the system would have identified decreases both the numerator and the denominator, which can lower recall, whereas removing a true pair that the system would not have identified decreases only the denominator, which can raise recall.

In both experiments, SHA-256 achieves the highest F-score, followed by DINO and ResNet at comparable levels, then CLIP and pHash in turn. By design, SHA-256 hashes are statistically independent across all content, thereby strictly limiting illegitimate pairs above the threshold. In contrast, pHash captures perceptual similarity and assigns high scores to visually related content from a common lineage. This behaviour increases the number of claimed pairs, depressing precision despite its robustness elsewhere.

\section{Discussion}
The following discussion considers what the present work leaves unresolved and where future research may be directed.

\subsection{On cooperative and adversarial environments}
The proposed scheme presupposes the cooperation of the generative platform. It is only within such a cooperative framework that steganographic inheritance takes place at each act of generation. Once content departs that framework and is regenerated by external tools or adversarial parties, the inheritance chain is severed at that point, and subsequent descendants may carry no traceable trait of their true lineage. In this sense, the scheme does not claim immunity to adversarial environments, and passive forensic inference may remain a necessary complement where cooperation cannot be assumed. It is worth noting, however, that the system was evaluated not only under common processing operations but also under semantic re-generation by generative models, and the results, whilst imperfect, were by no means without promise. The space of possible transformations is, of course, inexhaustible, and no training regime can anticipate every condition a piece of synthetic content may encounter in the wild. Whether the robustness of steganographic inheritance can be extended systematically to a broader and more adversarial range of re-generative processes remains an open question.

\subsection{On projectors and stegosystems}
The present study evaluates a wide range of projectors, each of which demonstrates serviceable performance within its respective domain of strength. Yet none was designed with the specific purposes of phylogenetic inference in mind. The development of projectors optimised for this setting represents a potential direction for future enquiry. One of the instructive findings of this study is that classical stegosystems based upon the principles of signal processing, conceived decades before the era of generative AI, exhibit a non-negligible degree of robustness to such transformations. This serves as a reminder that efficacy is not determined by recency, and that the insights accumulated over decades of signal processing research need not be set aside in favour of data-driven methods alone. As capacity grows, whether through architectural advances or the principled integration of classical techniques, richer representations of lineage become possible, accommodating more complex genealogical information within a single act of inheritance.

\subsection{On cross-modalities}
While the implementation presented in this study is conducted on digital imagery, the underlying framework of steganographic inheritance is not necessarily confined to this modality. Synthetic text, audio and video are equally subject to the genealogical questions this study seeks to address, as well as the cross-modal relations between them. A trait embedded in an image might be matched against the representation of a text or an audio signal from which it descends, or vice versa. This remains a largely open frontier.

\subsection{On multi-generational phylogeny}
The present system is designed to recover the immediate parent-offspring pair and to answer the question of direct descent. From successive matchings, a complete phylogenetic tree may be reconstructed, provided no intermediate nodes are missing. However, the reconstruction of multi-generational relationships from a single descendant alone remains beyond its present scope. One natural direction is to explore the co-existence of multi-generational trait information within a single piece of media, allowing it to carry forward not only the connection to its immediate parent but traces of more distant ancestors as well. Whether a more elegant mechanism may be devised to this end invites further investigation.

\subsection{On multidisciplinary collaboration}
The technical framework proposed in this study may not be a sufficient condition for the practical governance of synthetic information lineage. For the scheme to operate within the real contexts of law, journalism and creative industries, it must be accompanied by engagement with scholars and practitioners across disciplines. Whether a platform bears an ethical obligation to maintain the inheritance chain of content it generates, and what evidentiary standing an extracted trait may carry, are questions that perhaps exceed the scope of pure technology, but that will ultimately determine whether the system described here serves the purposes for which it is intended.

\section{Conclusion}

This study has proposed steganographic inheritance as a mechanism for tracing the phylogeny of synthetic information, embedding the lineage of digital content at the moment of its creation so that its origin may be recovered long after the ancestral form has passed beyond reach. A theoretical framework characterises the conditions under which such recovery is reliable, and empirical evaluations across a range of projectors, stegosystems and transformation conditions demonstrate that the scheme is not merely conceivable but practicable. Darwin closed his great work, \emph{On the Origin of Species}, with a reflection that has lost none of its resonance across the span of history~\cite{darwin1859}:
\begin{quote}
\emph{There is grandeur in this view of life, with its several powers, having been originally breathed into a few forms or into one; and that, whilst this planet has gone cycling on according to the fixed law of gravity, from so simple a beginning endless forms most beautiful and most wonderful have been, and are being, evolved.}
\end{quote}
We venture a counterpart vision for the information age. There is awe in this view of synthetic information, with its several manifestations, having originally emerged from a few generative models; and that, whilst AI has gone marching on according to the neural scaling law, from so simple a beginning endless forms most lifelike and most consequential have been, and are being, evolved.

\section*{Acknowledgements}
This work was supported in part by the Japan Society for the Promotion of Science (JSPS) under KAKENHI Grants JP21H04907 and JP24H00732, and in part by the Japan Science and Technology Agency (JST) under the CREST Grants JPMJCR20D3 and JPMJCR2562, including the AIP Challenge Program, and under the AIP Acceleration Grant JPMJCR24U3 and under the K Program Grant JPMJKP24C2.

\IEEEtriggeratref{109}
\bibliography{Transactions-Bibliography/bstcontrol, Bib/bib_tree}
\bibliographystyle{Transactions-Bibliography/IEEEtran}

\end{document}